\documentclass{IEEEtran}
\usepackage{amsmath,epsfig}
\usepackage{amssymb,mathrsfs}
\usepackage{graphicx}
\usepackage{epsfig}
\usepackage{hyperref}
\usepackage{multirow}
\usepackage{subfigure}
\usepackage{pstool}
\usepackage{cite}

\usepackage{algorithm}
\usepackage{algorithmic}

\usepackage[switch]{lineno}

\usepackage{amssymb}
\usepackage[mathscr]{euscript}

\title{Deep Learning and Music Adversaries}

\author{Corey Kereliuk,~\IEEEmembership{Member,~IEEE,}
Bob L. Sturm,~\IEEEmembership{Member,~IEEE,} 
Jan Larsen~\IEEEmembership{Senior Member,~IEEE}
\thanks{C. Kereliuk and J. Larsen are with DTU Compute, Technical University of Denmark.}
\thanks{B. L. Sturm is with the School of
Electronic Engineering and Computer Science, 
Queen Mary University of London.
}}%

\begin{document}
\maketitle

\begin{abstract}
An {\em adversary} is essentially an algorithm 
intent on making a classification system perform in 
some particular way given an input, e.g., 
increase the probability of a false negative.
Recent work builds adversaries for deep learning 
systems applied to image object recognition,
which exploits the parameters of the system
to find the minimal perturbation of the input image
such that the network misclassifies it with high confidence.
We adapt this approach to construct and deploy an adversary of 
deep learning systems applied to music content analysis.
In our case, however, the input to the systems 
is magnitude spectral frames,
which requires special care in order to produce valid input audio signals
from network-derived perturbations.
For two different train-test partitionings of two benchmark datasets, 
and two different deep architectures,
we find that this adversary is very effective in defeating the resulting systems.
We find the convolutional networks are more robust, however,
compared with systems based on a majority vote over individually classified audio frames.
Furthermore, we integrate the adversary into the training of new deep systems,
but do not find that this improves their resilience against the same adversary.
\end{abstract}


\section{Introduction}
Deep learning is impacting the research domain of
music content analysis and music information retrieval (MIR)\cite{Lee2009c,Hamel2010,Yang2011b,Humphrey2013,Oord2013,
Pikrakis2013,Sigtia2014,Dieleman2014,Zhang2014},
but recent developments raise the spectre that 
the high performance of these systems 
does not reflect how well they have learned to solve high-level problems
of music listening.
MIR aims to produce systems that help make 
``music, or information about music, easier to find'' \cite{Casey2008a}. 
This is of principal importance for
confronting the vast amount of music data
that exists and continues to be created. 
Listening machines that can flexibly produce accurate, 
meaningful and searchable descriptions of music 
can greatly reduce the cost of processing music data,
and can facilitate a diversity of applications. 
These extend from music identification \cite{Wang2003}, 
author attribution \cite{Casey2008}, recommendation \cite{Oord2013},
transcription \cite{Ewert2014},
and playlist generation \cite{Aucouturier2002b},
to extracting semantic descriptors such as genre and mood 
\cite{Bertin-Mahieux2010c,Yang2011f,Sturm2013h}, 
to computational musicology \cite{Collins2010}, 
and even synthesis and music composition \cite{Schwarz2006}.

Recent surveys of the domain of deep learning 
record impressive results for several benchmark problems \cite{Deng2014,Bengio2015a}.
In addition to these major successes,
deep learning methods are very attractive for three other reasons:
there now exist efficient and effective training algorithms for deep learning,
not to mention completely free and open cross-platform implementations, 
e.g., {\em Theano} \cite{Bastien2012,Bergstra2010};
they entail jointly optimising feature learning and classification,
thus allowing one to forgo many difficulties inherent to 
formally encoding expert knowledge into a machine;
and their layered structures seems to favour hierarchical representations
of structures in data.
One caveat, however, is that these methods require a lot of data
in order to estimate parameters and generalise well \cite{Montavon2012}.

In MIR, the works in \cite{Lee2009c,Hamel2010,Li2010}
are among the first to apply deep learning to music content analysis,
and each describes results pointing to the conclusion that
these systems can automatically learn features relevant for
complex music listening tasks, e.g., recognition of genre or style.
Results since then point to the same conclusion 
\cite{Yang2011b,Pikrakis2013,Sigtia2014,Dieleman2014,Zhang2014}.
Humphrey et al. \cite{Humphrey2013} highlight this fact
to argue deep learning is naturally suited to learn 
relevant abstractions for music content analysis,
provided enough data is available.
Since music can be seen as a ``whole greater than the sum of its parts'' \cite{Humphrey2013},
deep learning can help MIR narrow the ``semantic gap'' \cite{Wiggins2009},
and move beyond what has been called a ``glass ceiling'' in performance \cite{Aucouturier2004}.

However, it is now known how deceiving the appearance
of high performance can be:
an MIR system can appear to be very successful in solving a high-level
music listening problem when in fact it is just exploiting
some independent variables of questionable relevance 
unknowingly confounded 
with the ground truth of a music dataset
by a poor experimental design 
\cite{Pampalk2005b,Flexer2007,Flexer2010,Sturm2012e,Urbano2013,Sturm2013h,
Sturm2013g,Sturm2013h,Sturm2015a,Sturm2015b,Sturm2014}.
In addition, recent work in machine learning has demonstrated
deep learning systems behaving in ways that contradict their appearance of
solving content-recognition problems.
Nguyen et al. \cite{Nguyen2014a} show how 
a high-performing image object recognition system 
can label with high confidence non-sensical synthetic images.
In a similar direction, we have shown \cite{Sturm2015b}
how a deep system that appears highly capable of recognising different
musical rhythms confidently classifies synthesised rhythms, 
though they bear little similarity to the rhythms they supposedly represent.
Szegedy et al. \cite{Szegedy2014} 
show how deep high-performing image object recognition systems
are highly sensitive to imperceptible perturbations
created by an {\em adversary}:
an agent that actively seeks to fool a classifier 
by perturbing the input such that it results in 
an incorrect output but with high confidence \cite{Dalvi2004a}.

All of these results 
motivate several timely questions of deep learning 
systems for music content analysis specifically,
and multimedia in general.
First, how do the adversaries of Szegedy et al. \cite{Szegedy2014} 
translate to the context of deep learning applied to music content analysis?
The input of the systems studied by Szegedy et al. \cite{Szegedy2014} 
is raw pixel data; however, in music content analysis
only the system studied in \cite{Dieleman2014} 
takes as input raw audio samples.
The inputs to other deep learning systems have been features:
windowed magnitude spectra \cite{Hamel2010,Sigtia2014},
sonograms \cite{Lee2009c,Oord2013},
autocorrelations of spectral energies \cite{Pikrakis2013,Sturm2015b},
or statistics of features \cite{Yang2011b,Zhang2014}.
Second, can we generate an adversary for such 
deep learning music content analysis systems 
that produce adversarial examples that are perceptually identical 
to the originals?
Third, can we ``harness'' an adversary to train 
deep learning systems that are robust to its ``malfeasance''?
Finally, and more broadly, what is deep learning 
contributing to music content analysis?
Can we use adversaries to reveal whether 
these deep systems are using better models of the content
than other state of the art systems
using hand-crafted features?

Our preliminary work \cite{Kereliuk2015}
shows that it is possible to create highly effective adversaries
of the music content analysis deep neural networks (DNN)  
studied in \cite{Hamel2010,Sigtia2014}.
These adversaries can make the systems
always wrong, always right, and anywhere in-between,
with high confidence by applying only minor perturbations
of the input magnitude spectra.
Furthermore, we created an ensemble of adversaries 
that can coax the DNN into assigning with high confidence
any label to the same music by perturbing 
the input by very small amounts (e.g., $26.8$ dB SNR).
In this article, we greatly expand upon our prior work \cite{Kereliuk2015}
to include convolutional deep learning systems,
more extensive testing in a larger benchmark MIR dataset,
and the results of incorporating an adversary into the training
of these different deep learning systems.

In the next section, we provide an overview of work 
applying deep learning to music content analysis and MIR.
We then review two different deep learning architectures,
and our construction of several music content analysis systems
using two partitions of two MIR benchmark datasets.
In Sec.~\ref{sec:adversaries} we review adversaries,
and design an adversary for our deep systems.
We then present in Sec. \ref{sec:experimental_results} a series of 
experiments using our adversary.
In Section V we provide a discussion of our work in wider contexts.
We conclude in section VI.
Some of our results can be produced with the software here:
\url{https://github.com/coreyker/dnn-mgr}.

\section{Deep Learning for Music Content Analysis}
\label{sec:review}
We first provide an overview of research in applying 
deep learning approaches to music content analysis.
We then discuss two different architectures, 
train two music content analysis systems,
and test them in two benchmark MIR datasets.
These systems are the subjects of our experiments 
in Section~\ref{sec:experimental_results}.

\subsection{Overview}
Artificial neural networks have been applied 
to many music content analysis problems, \cite{Griffith1999}, 
for instance, fingerprinting \cite{Burges2003},
genre recognition \cite{Matityaho1995},
emotion recognition \cite{Vempala2012},
artist recognition \cite{Whitman2001},
and even composition \cite{Papadopoulos1999a}.
Advances in training have enabled the creation of 
more advanced and deeper architectures.
Deng and Yu \cite{Deng2014} (Chapter 7) 
provide a review of successful applications of  
deep learning to the analysis of audio,
highlighting in particular its significant contributions 
to speech recognition in conversational settings.
Humphrey et al. \cite{Humphrey2013} 
provide a review for applications to music in particular,
and motivate the capacity of deep architectures to automatically learn 
hierarchical relationships in accordance
with the hierarchical nature of music: ``pitch and loudness combine 
over time to form chords, melodies and rhythms.''
They argue that this is key for moving beyond
the reliance on ``shallow'' and hand-designed 
features that were designed for different tasks.

Lee et al. \cite{Lee2009c} are perhaps the first to apply 
deep learning to music content analysis, 
specifically genre and artist recognition.
They train a convolutional deep belief network (CDBN) with two hidden layers
in an unsupervised manner in an attempt to make the hidden layer activations produce meaningful features from a pre-processed spectrogram input
computed using 20 ms 50\%-overlapped windows.
The spectrogram is ``PCA-whitened'', which involves projecting it
onto a lower-dimensional space using scaled eigenvectors.
Important details are missing in the description of the work,
but it appears they use the activations as features
in some train/test task using a standard machine learning approach.
A table of their experimental results, using some portion of the dataset {\em ISMIR2004},
shows higher accuracies for their deep learned features compared to those for standard MFCCs.
For genre recognition,
Li et al. \cite{Li2010} use convolutional deep neural networks (CDNN) with three hidden layers,
into which they input a sequence of 190 13-dimensional MFCC feature vectors.
The architecture of their CDNN is such that the first hidden layer 
considers data from 127 ms duration, and the last hidden layer 
is capable of summarising events over a 2.2 s duration.
van den Oord et al. \cite{Oord2013} apply CDNN to mel-frequency spectrograms 
for automatic music content analysis.

For genre recognition and more general descriptors,
Hamel and Eck \cite{Hamel2010} train a DNN
with three hidden layers of 50 units each,
taking as input 513 discrete Fourier transform (DFT) 
magnitudes computed from a single 46 ms audio frame.
They use a train/valid/test partition 
of the benchmark music genre dataset {\em GTZAN} \cite{Tzanetakis2002,Sturm2013h}.
They also explore ``aggregated'' features, which 
are the mean and variance in each dimension of activations 
over 5 second durations.
They find in the test set, and for both short-term and aggregated features,
that SVM classifiers trained with features built from hidden layer activations 
reproduce more ground truth than an SVM classifier
trained with features built from MFCCs.
They report an accuracy of over 0.84 for features 
that aggregate activations of all three hidden layers.
Sigtia and Dixon \cite{Sigtia2014} explore modifications
to the system in \cite{Hamel2010}, in particular
using different combinations of architectures, 
training procedures, and regularisation.
They use the activations of their trained DNN as features
for a train/test task using a random forest classifier. 
They report an accuracy of about 0.83 using features 
aggregating activations of all hidden layers of 500 units each.
For genre recognition,
Yang et al. \cite{Yang2011b} combine 263-dimensional modulation features with a DBN.
For music rhythm classification,
Pikrakis \cite{Pikrakis2013} employs a DBN, 
which we studied further in \cite{Sturm2014,Sturm2015a,Sturm2015b}.

\begin{figure*}[ht]
\centering
\includegraphics[width=0.95\linewidth]{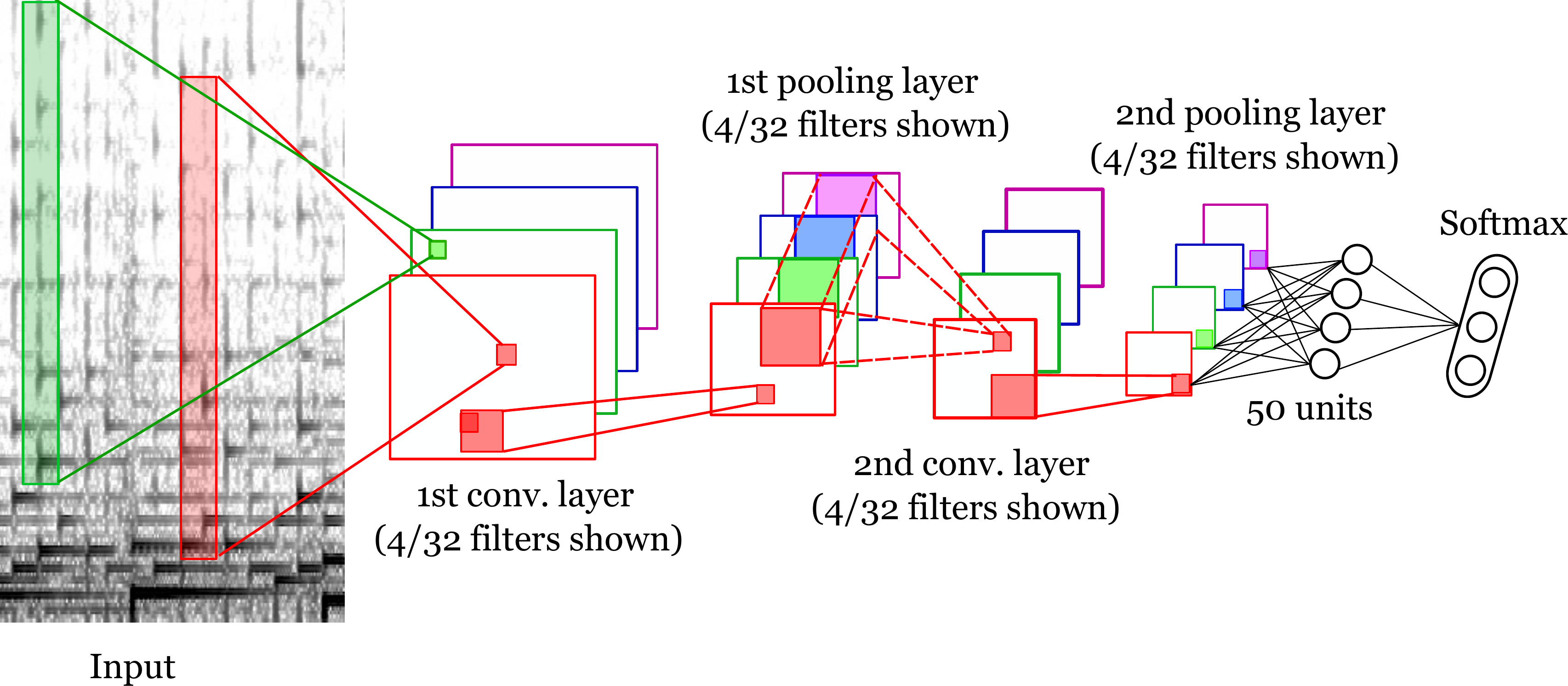}
\caption{Illustration of the CDNN architecture we use for our experiments. 
The CDNN first applies narrow vertical filters to the input sonogram (left) 
to capture harmonic structure. 
Then, it applies 32 different filters in the first convolutional layer (we show only 4). 
This is followed by the first max-pooling layer,
and then a 2nd pair of convolutional and max-pooling layers.
Finally, the output of the final max-pooling layer
is fully connected to a final hidden layer of 50 units, 
followed by a softmax output unit. 
The input spectrogram contains 100 time slices, 
which means that the final layer of the CDNN summarises 
information over a total duration of 2.35 seconds.}
\label{fig:CDNN}\vspace{-0.1in}
\end{figure*}

Dieleman et al. \cite{Dieleman2011} build and apply CDBN
to music key detection, artist recognition, and genre recognition.
There are three major differences with respect to the work above 
\cite{Lee2009c,Yang2011b,Hamel2010,Li2010,Sigtia2014}.
First, Dieleman et al. employ 24-dimensional input features computed by 
averaging short-time chroma and timbre features over the 
time scales of single musical beats.
Second, they employ expert musical knowledge
to guide decisions about the architecture of the system.
Finally, they use the output posteriors of their system for classification, 
instead of using the hidden layer activations as features for a separate classifier.
Their experiments in a portion of the ``million song dataset'' \cite{Bertin-Mahieux2011}
show large differences in classification accuracies
between their systems and a naive Bayesian classifier using the 
same input features.
In a unique direction for audio,
Dieleman and Schrauwen \cite{Dieleman2014} explore
``end-to-end'' learning, where a CDNN is trained with input of
about 3 s of raw audio samples for a music content analysis task (autotagging).
They find that the lowest layer of the trained CDNN
appears to learn some filters that are frequency selective.
They evaluate this system for a multilabel problem.

To recognise music mood,
Weninger et al. \cite{Weninger2014} use recurrent DNN
with input constructed of several statistics of 
low-level features computed over second-long excerpts of music recordings.
Battenberg and Wessel \cite{Battenberg2012} apply DBN for identifying the beat numbers 
over several measures of percussive music,
with input features consisting of quantised onset times and magnitudes.
Boulanger-Lewandowski et al. \cite{Boulanger-Lewandowski2013} 
train a recurrent neural network to produce chord classifications
using input of PCA-whitened magnitude DFT.
In a similar direction, Humphrey and Bello
\cite{Humphrey2014} build a DNN that maps
input spectrogram features to guitar-specific 
fingerings of chords.


\subsection{Two types of deep architectures}
\label{sec:dnn-systems}
We now review two different architectures of deep learning systems,
and the way they are trained.
A DNN is an artificial neural network with several hidden layers \cite{Deng2014}. 
The output of each layer is a non-linear function of its inputs, 
obtained by a matrix multiplication cascaded with a non-linearity, 
e.g., tanh, sigmoid and rectifier. 
By chaining together several hidden layers, 
composite representations of the input emerge in deeper layers. 
This fact can give deep networks greater representational power 
than shallower networks containing an equivalent number of parameters \cite{Bengio2009}.

A CDNN is a special type of DNN with weights 
that are shared between multiple points between adjacent layers. 
The weight sharing in CDNNs not only reduces the number of trainable parameters,
but also causes matrix multiplications to reduce to convolutions,
which can be implemented efficiently. 
Furthermore, many natural signals have local spatial or temporal structures that are repeated globally.
For example, natural images often consist of oriented edges;
and audio signals often consist of harmonic and repetitive structures.
CDNNs can learn these types of structures very well.
Figure \ref{fig:CDNN} illustrates our CDNN,
which we discuss in the following subsection.

The contemporary success of deep learning 
comes with computationally efficient training methods.
Systems that have such deep architectures 
are usually trained using gradient descent, which consists of 
backpropagating error derivatives from the cost function through the network. 
There are a plethora of useful tips and tricks to augment training,
including stochastic gradient descent, dropout regularisation, 
weight decay, momentum, learning rate decay, and so on \cite{Montavon2012}.

\subsection{Deep learning with two music genre benchmarks}\label{sec:oursystems}
We now build DNNs and CDNNs using two music genre benchmarks:
{\em GTZAN} \cite{Sturm2013h,Tzanetakis2002} 
and the Latin Music Database ({\em LMD}) \cite{Silla2008b}.
{\em GTZAN} consists of 100 30-second music recording excerpts in each of ten categories,
and is the most-used public dataset in MIR research \cite{Sturm2014d}.
{\em LMD} is a private dataset, 
consisting of 3,229 full-length music track recordings
non-uniformly distributed among ten categories,
and has been used in the annual MIREX audio latin music genre classification
evaluation campaign since 2008.\footnote{\url{http://www.music-ir.org/mirex/wiki/MIREX_HOME}}
We use the first 30 seconds of each track in {\em LMD}.

We build several DNNs and CDNNs using different partitionings of these datasets.
One partitioning of {\em GTZAN} we create by randomly selecting
500/250/250 excerpts for training/validation/testing.
The other partitioning of {\em GTZAN} is ``fault-filtered,'' 
which we construct by hand to include 443/197/290 excerpts.
This involves removing $70$ files including exact replicas, 
recording replicas, and distorted files \cite{Sturm2013h},
and then dividing the excerpts such that no artist is repeated 
across the training, validation, and test partitions. 
We partition  {\em LMD} in two ways:
1) partitioning by 60/20/20\% sampling in each class;
2) a hand-constructed artist-filtered partitioning
containing approximately the same division of excerpts
in each class.
We retain all 213 replicas in {\em LMD}.\footnote{\scriptsize\url{https://highnoongmt.wordpress.com/2014/02/08/faults_in_the_latin_music_database}}

The input to our systems is derived from the short-time Fourier transform (STFT) 
of a sampled audio signal $x$ \cite{Allen1977a}:
\begin{align}
\mathcal{F}(x)[m,u] &= \sum_{l=0}^{L-1} w[l]x[l-uH]e^{-j2\pi m l/L}
\label{eq:dft}
\end{align}
where the parameter $L$ defines both the window length and the number of frequency bins. 
We define $w$ as a Hann window of length $L=1024$, which corresponds to a duration of 46ms
for recordings sampled at 22050 Hz. The window is hopped along $x$ with a stride of $H=512$ samples 
(adjacent windows overlap by 50\%). 

Since audio signals can be of any duration, we define the input to our systems as a sequence $X = (X_n)_{n=0}^{N-1}$, where the sequence length depends on the input audio's duration. We define the n$th$ element of the input sequence $X$ to be 
\begin{equation}
X_{n} \triangleq \Big( \Big| \mathcal{F}(x)[m,u] \Big| : m \in [0,512], u \in [nT, (n+1)T[ \Big)
\label{eq:in_seq}
\end{equation}
where $T=1$ for each DNN and $T=100$ for each CDNN. 
Thus, when $T=1$, $X$ is a sequence of $513\times 1$ vectors; 
when $T=100$, $X$ is a sequence of $513\times100$ matrices.

A (C)DNN processes each element in this sequence independently, outputting a sequence $P = (P_n)_{n=0}^{N-1}$ from the final (softmax) layer. The output vector $P_n \in [0,1]^K$, $\|P_n\|_1 = 1$, is the posterior distribution of labels assigned to the n$th$ element in the input sequence by the network. Therefore, we may write \mbox{$P_n(i|X_n, \Theta) \equiv P_n(i) \in [0,1]$} where $\Theta$ represents the trainable network parameters, i.e., the set of weights and biases. 
We define the {\em confidence} of a (C)DNN in a particular label $k\in \{1,\ldots,K\}$
for an input sequence $X$ as the sum of all posteriors, i.e.,
\begin{equation}
R(k|X,\Theta) = \frac{1}{N}\sum_{n=0}^{N-1} P_n(k|X_n, \Theta).
\label{eq:confidence}
\end{equation}
We apply a label to an input sequence $X$ 
as the one maximising the confidence
\begin{equation}
y(X,\Theta) = \arg\max_{k \in \{1,\ldots,K\}} R(k|X,\Theta).
\end{equation}


Paralleling the work in \cite{Sigtia2014}, we build DNNs with 3 fully connected hidden layers, and either 50 or 500 units per layer. Our CDNN has two convolutional layers (accompanied by max pooling layers) followed by a fully connected hidden layer with 50 units. Figure \ref{fig:CDNN} illustrates the architecture of our CDNN. Its first convolutional layer contains 32 filters, each arranged in a rectangular $400\times 4$ grid. We choose this long rectangular shape instead of the small square patches typically used when training on images based on our knowledge that many sounds exhibit strong harmonic structures that span a large portion of the audible spectrum. The second convolutional layer contains 32 filters, each connected in an $8\times8$ pattern. Our two pooling neighborhoods are $4\times4$ and have strides of $2\times2$.  
All of our deep learning systems use rectified linear units (ReLUs),
and have a softmax unit in the final layer.
As is typical, we standardise the (C)DNN inputs by subtracting the training set mean and 
dividing by the standard deviation in each of the input dimensions. 
We perform this with a linear layer above the input layer of each network.
The raw inputs to the network are still $X_n$.

\begin{figure*}[t]
\centering
\subfigure[Random partitioning]{
\includegraphics[width=0.45\linewidth]{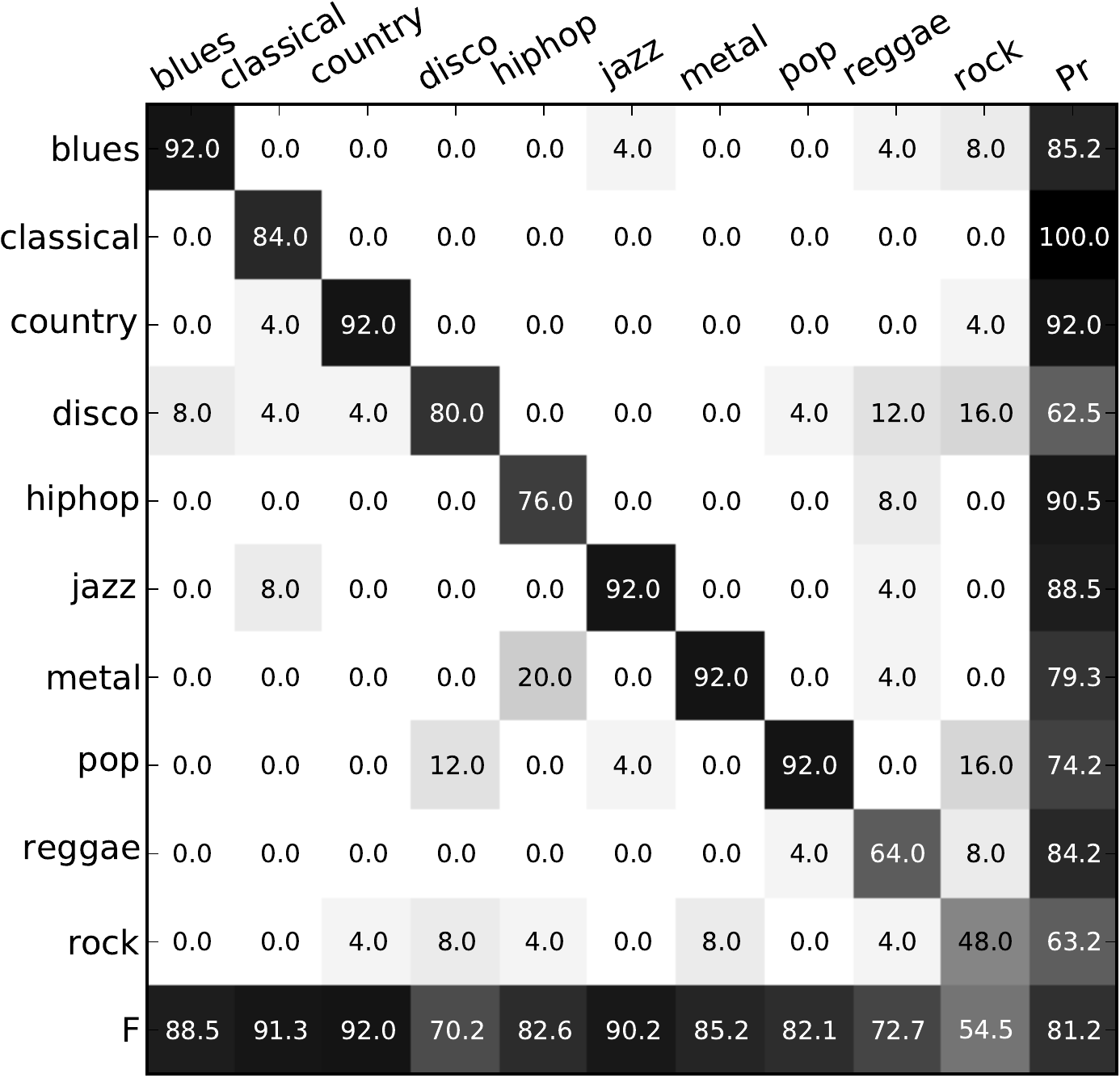}}
\subfigure[Artist-filtered partitioning]{
\includegraphics[width=0.45\linewidth]{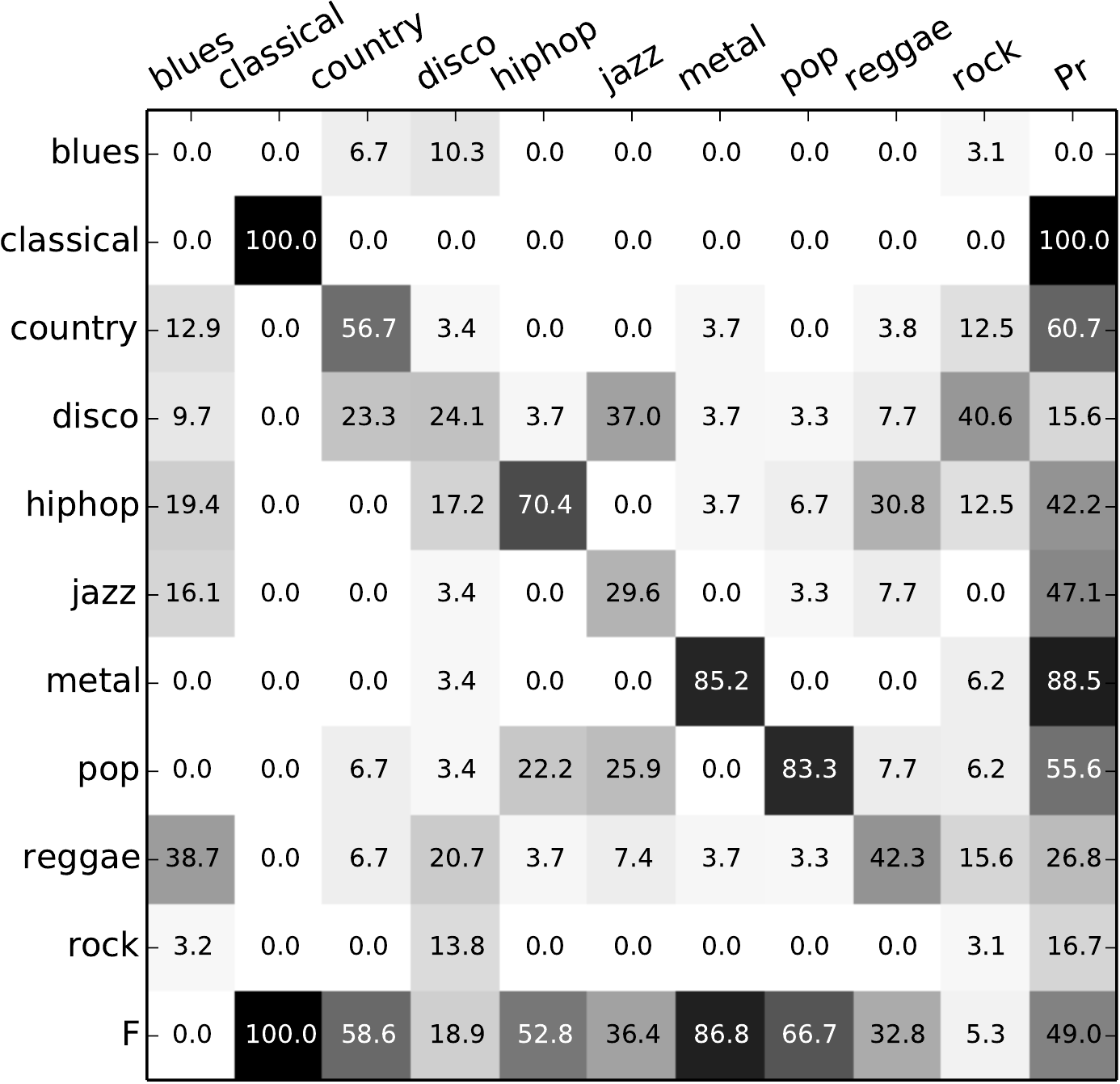}}
\caption{Figure of merit (FoM, $\times 100$) in {\em GTZAN}
with two different partitionings for random forest classification (majority vote)
of DNN-based features (all layers) aggregated over 5 second windows
(mean and standard deviations).
Each DNN has 500 rectified linear units in each hidden layer.
Columns represent the true class; rows denote labels chosen by system; the diagonal contains the per-class recall; the off-diagonal entries are confusions; the rightmost column is the precision; the bottom row is the F-score; and the last element along the diagonal is the mean recall (normalised classification accuracy).}
\label{fig:DNN-GTZAN-baselines}
\end{figure*}


Also paralleling \cite{Sigtia2014}, we build several music classification systems 
treating our DNN as a feature extractor.
In this case, we construct a set of features 
by concatenating the activations from the DNN's three hidden layers,
and aggregating them over 5-second texture windows (hopped by 50\%).
The aggregation summarises the mean and standard deviation of the feature dimensions 
over the texture window and may be seen as a form of late-integration of temporal information. We use this new set of features to train a random forest (RF) classifier \cite{Hastie2009} with 500 trees.
Thus, to classify a music audio recording $x$ 
from its set of aggregated features,
we use majority voting over all classifications,
which is also used in \cite{Sigtia2014}.

\subsection{Preliminary evaluation}
\begin{figure*}[ht]
\centering
\subfigure[DNN Random partitioning]{
\includegraphics[width=0.45\linewidth]{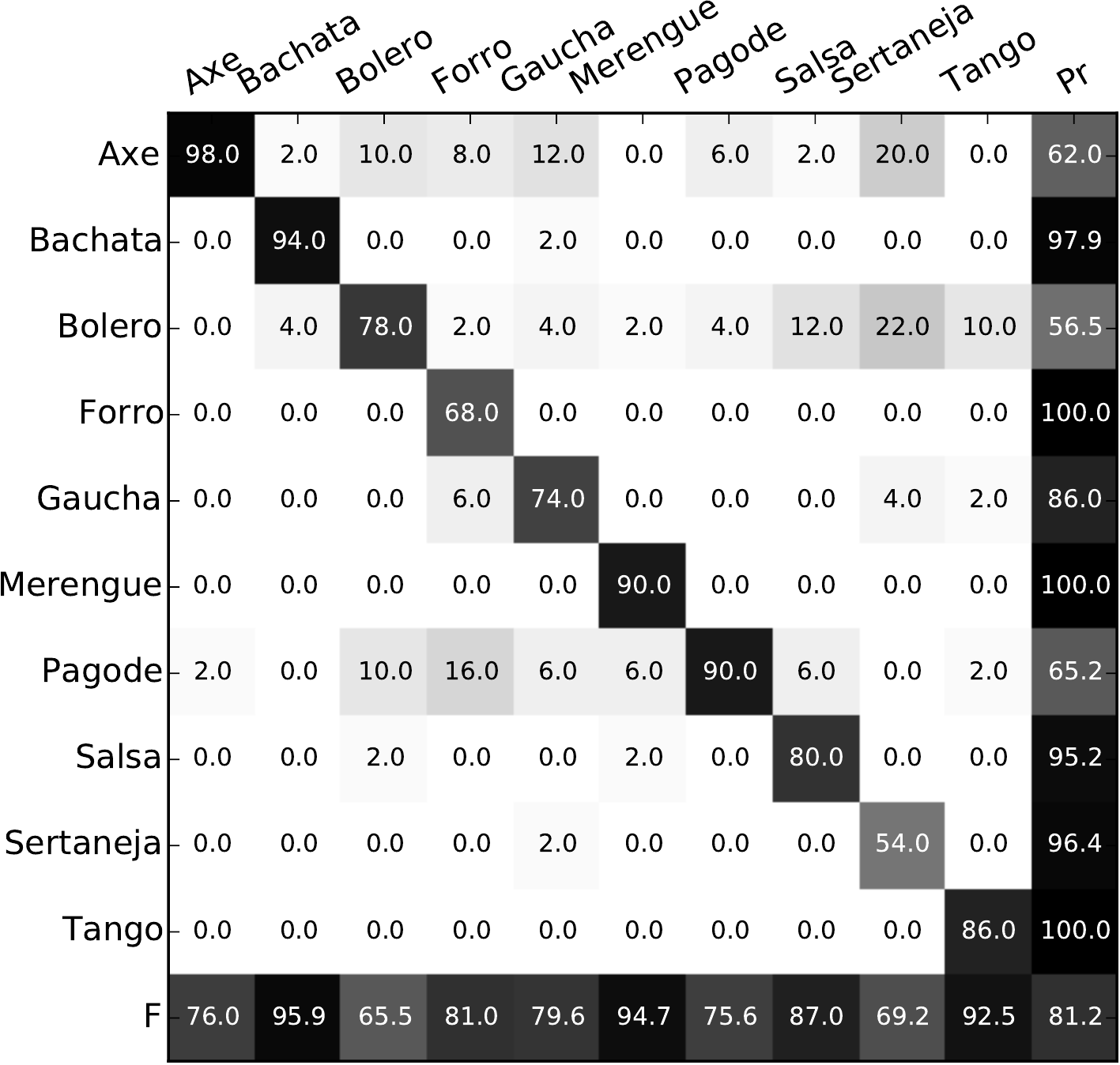}}
\subfigure[DNN Artist-filtered partitioning]{
\includegraphics[width=0.45\linewidth]{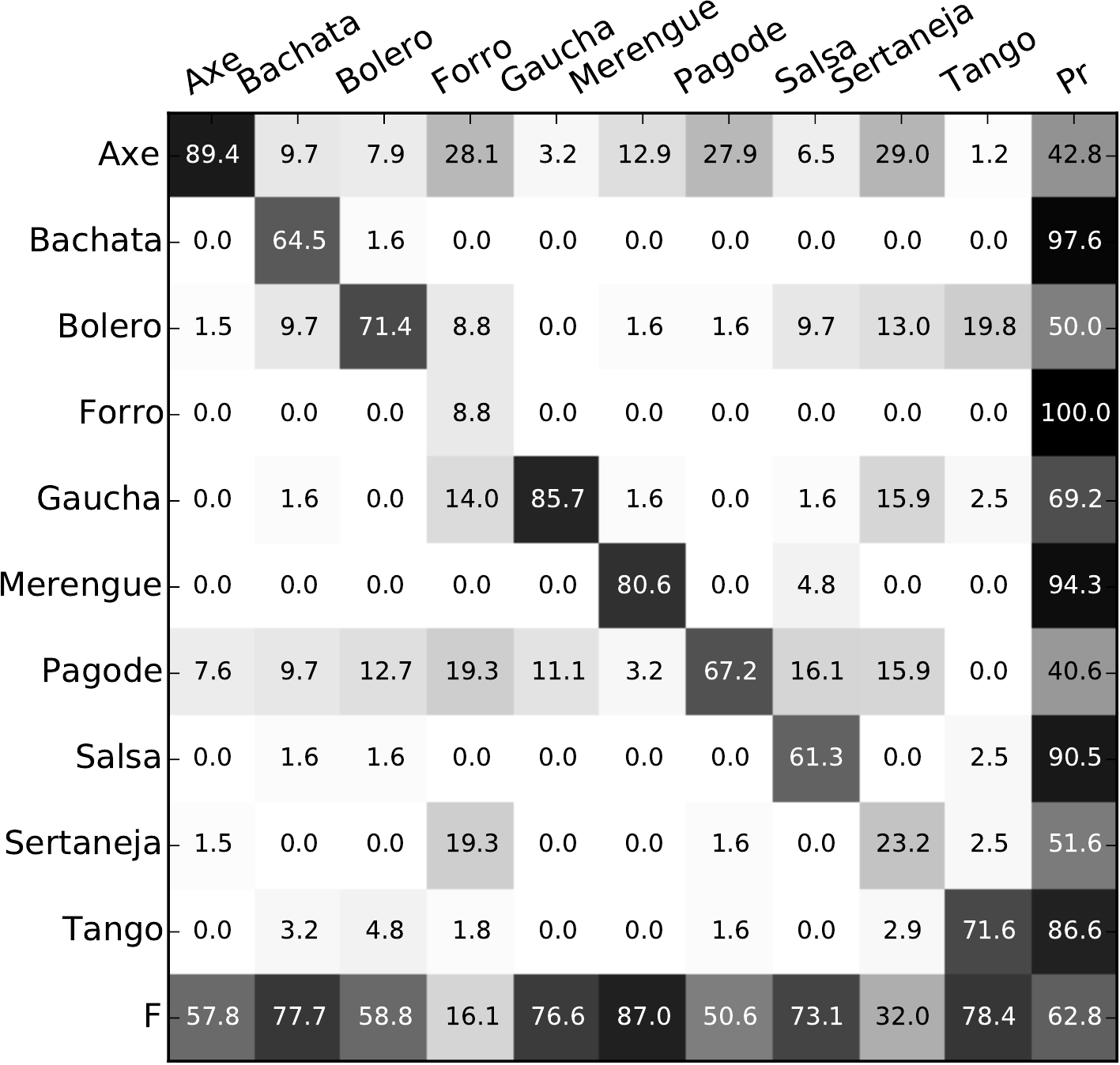}}\\
\subfigure[CDNN Random partitioning]{
\includegraphics[width=0.45\linewidth]{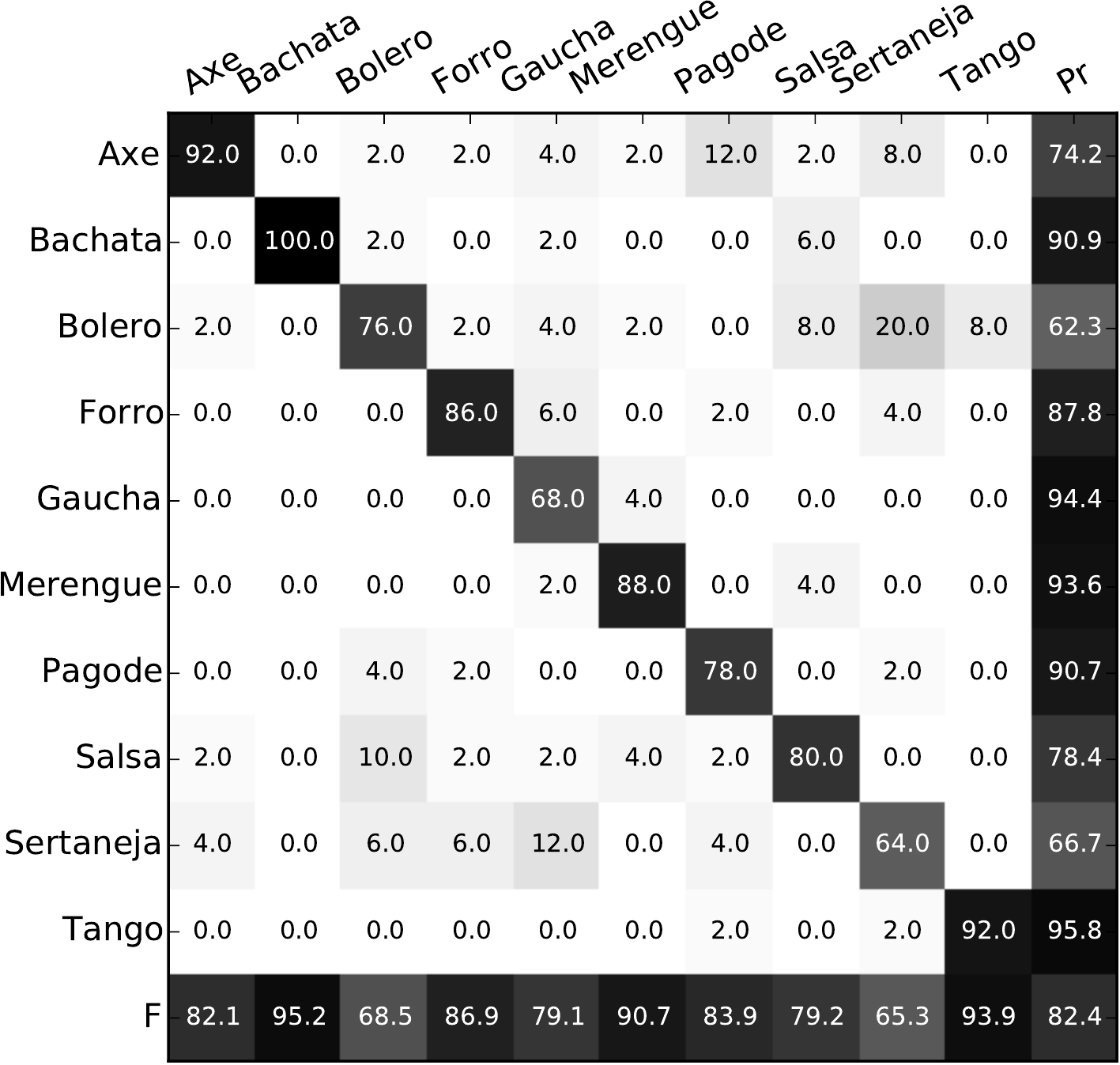}}
\subfigure[CDNN Artist-filtered partitioning]{
\includegraphics[width=0.45\linewidth]{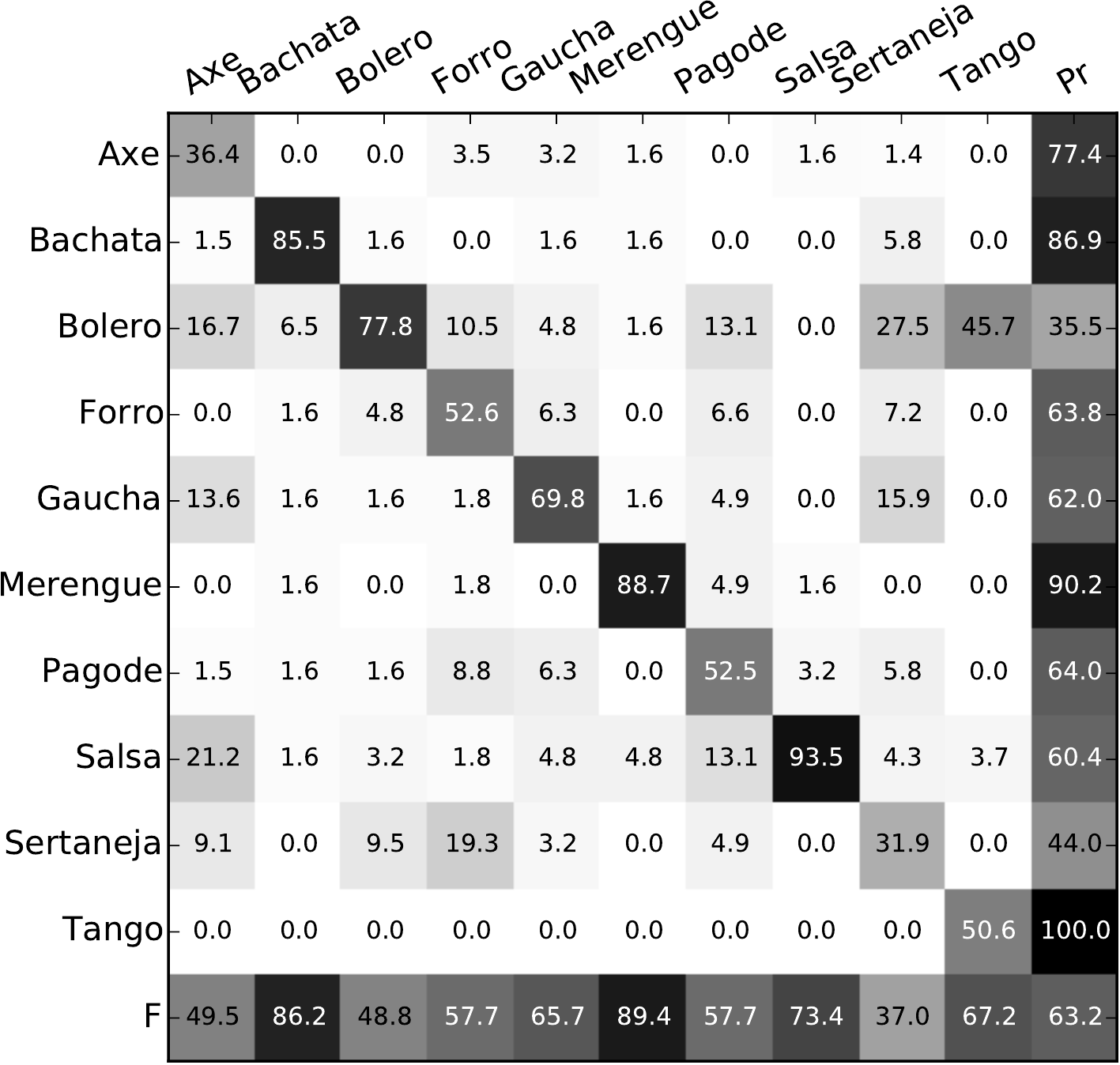}}
\caption{FoM for deep learning systems with two different partitioning strategies of {\em LMD}.
Interpretation as in Fig. \ref{fig:DNN-GTZAN-baselines};
but note that in this case we are using the deep learning systems 
as the classifiers, instead of performing classification 
using a random forest with features derived from hidden layer activations.
}
\label{fig:deep-LMD-baselines}\vspace{-0.1in}
\end{figure*}

Figure~\ref{fig:DNN-GTZAN-baselines} and Table~\ref{tab:SigtiaTable}
show the results of RF classification 
using the features produced by the DNN when trained on {\em GTZAN}
with the two different partitioning strategies;
and Fig. \ref{fig:deep-LMD-baselines} shows those for 
the (C)DNNs we train and test in {\em LMD}.
Across each partition strategy we see significant differences in performance.
The mean recall in each class in Figure~\ref{fig:DNN-GTZAN-baselines}
on the fault-filtered partition is much lower than that on the random test partition ---
involving drops higher than 30 percentage points in most cases.
Table~\ref{tab:SigtiaTable} shows similar drops in performance
that persist over the inclusion of drop-out regularisation. 
Such significant drops in performance from partitioning based on artists is not unusual, 
and has been studied before as a bias coming from 
the experimental design \cite{Pampalk2005b,Flexer2010,Sturm2013h}.
Partitioning a music genre recognition dataset along
artist lines has been recommended to avoid this bias \cite{Pampalk2005b,Flexer2010},
and is in fact used in several MIREX audio classification tasks.\footnote{\url{http://www.music-ir.org/mirex/wiki/MIREX_HOME}}
Experiments using {\em GTZAN} with fault-filtering partitioning
has not been used in many benchmark experiments with {\em GTZAN}
because its artist information has only recently been made available \cite{Sturm2013h}.



\section{Adversaries in music content analysis}
\label{sec:adversaries}
An adversary is an agent that tries 
to defeat a classification system in order to maximise its gain, 
e.g., SPAM detection.
Dalvi et al. \cite{Dalvi2004a} pose this problem
as a game between a classifier and adversary,
and analyse the strategies involved for an 
adversary with complete knowledge of the classification system,
and for a classifier to adapt to such an adversary. 
Szegedy et al. \cite{Szegedy2014} propose using adversaries
for testing the assumption that deep learning systems
are ``smooth classifiers,'' i.e., stable in their classification to
small perturbations around examples in the training data. 
They define an adversary of a classifier $f: \mathbb{R}^m \to \{1, \ldots, K\}$
as an algorithm using complete knowledge of the classifier
to perturb an observation $x\in \mathbb{R}^m$
such that $f(x+r) \ne f(x)$, where $r\in \mathbb{R}^m$ is some small perturbation.
Specifically, their adversary solves the constrained optimisation problem
for a given $k\in\{1, \ldots, K\}$:
\begin{align}
\min \|r\|_2 \; \text{subject to} \; f(x+r) = k.
\end{align}
For $k \ne f(x)$, Szegedy et al. \cite{Szegedy2014} employ a line search
along the direction of the loss function of the network
starting from $x$ until the classifier produces the requested class.
They find that adversarial examples of one classifier
can fool other classifiers trained on independent data;
hence, one need not have complete knowledge of a classifier 
in order to fool it.

\begin{table}[t]
\begin{center}
\begin{tabular}{|c | c | c | c |}
  \hline
  Hidden Units & Layer & ReLU & ReLU+Dropout \\
  \hline
  \multirow{4}{*}{50} 
  & 1   & 76.00 (40.69) & 80.40 (45.17) \\
  & 2   & 78.80 (45.17) & 80.40 (43.10) \\
  & 3   & 79.60 (43.79) & 78.80 (44.48)  \\
  & All & 80.40 (43.79) & 80.00 (43.79)  \\
  \hline
  \multirow{4}{*}{500} 
  & 1    & 68.40 (40.34)  & 75.60 (40.69) \\
  & 2    & 74.40 (40.69)  & 80.00 (50.34) \\
  & 3    & 77.60 (43.79)  & 79.20 (48.62) \\
  & All  & 76.00 (42.41)  & 81.20 (48.97) \\
  \hline
\end{tabular}
\end{center}
\caption{Mean normalised classification accuracy ($\times 100$) in {\em GTZAN}
for random forest classification of DNN-based features from layer shown
aggregated over 5-second windows.
Number outside brackets is from
random partition in Fig. \ref{fig:DNN-GTZAN-baselines}(a);
and that inside brackets is from
fault-filtered partition in Fig. \ref{fig:DNN-GTZAN-baselines}(b).
}
\label{tab:SigtiaTable}\vspace{-0.3in}
\end{table}%

Goodfellow et al. \cite{Goodfellow2015a}
provide an intuitive explanation of these adversaries:
even though the perturbations in each dimension 
might be small, their contribution to the magnitude of a projection 
grows linearly with input dimensionality.
With a deep neural network involving many such projections in each layer, 
a small perturbation at its high-dimensional input layer 
can create major consequences at the output layer.
Goodfellow et al. \cite{Goodfellow2015a} show that adversarial examples
can be easily generated by making the perturbation 
proportional to the sign of the partial derivative of the 
loss function used to train a particular network,
evaluated with the requested class.
They also find that the direction of perturbation 
is important, not necessarily its size.
Hence, it seems adversarial examples of one model will likely fool other models because 
they occur in large volumes in high-dimensional spaces.
This is also found by Gu and Rigazio \cite{Gu2014}.

As for Szegedy et al. \cite{Szegedy2014}, we are interested the robustness of
our deep learning music content analysis systems to an adversary.
Do these systems suffer just as dramatically as the 
image content recognition systems in \cite{Szegedy2014,Gu2014,Goodfellow2015a}?
In other words, can we find imperceptible perturbations of audio recordings,
yet make the systems produce any label with high confidence?
If so, can we adapt the training of the systems 
such that they become more robust?
In the next subsections, we define an adversary as an optimisation problem,
but with care of the fact that the input to our deep learning systems
are magnitude STFT \eqref{eq:in_seq}.
We then present an approach to 
integrate adversaries into the training of our systems.
We present our experimental results in Section \ref{sec:experimental_results}.

\subsection{Adversaries for music audio}
The explicit goal of our adversary 
is to perturb a music recording $x$  
such that a system will confidently classify it with some class 
$y\in\{1,\ldots,K\}$.
Specifically, we define the adversary
as the constrained optimisation problem:
\begin{equation}
  \hat{X}(y) = \arg\min_{Z\in{C(X)}} \sum_{n=0}^{N-1}\mathcal{L}(Z_n, y|\Theta)
  \label{eq:opt-adverse}
\end{equation}
where we define the feasible set of adversarial examples to input sequence $X$ as:
\begin{align}
  \mathcal{C}(X) = \Big\{ Z &= (Z_n)_{n=0}^{N-1}: \nonumber\\
  &\sqrt{\textstyle\sum_{n=0}^{N-1}\|Z_n-X_n\|_2^2} \leq N\epsilon(\text{SNR}) \Big\}
  \label{eq:feasset}
\end{align}
with the parameter
\begin{equation}
  \epsilon(\text{SNR}) = \frac{\frac{1}{N}\sqrt{\textstyle\sum_{n=0}^{N-1} \|X_n\|_2^2}}{10^{\text{SNR}/20}}
  \label{eq:snr}
\end{equation}
limiting the maximum acceptable perturbation caused by the adversary. 
The loss function in \eqref{eq:opt-adverse} is the 
cross-entropy loss function,
$\mathcal{L}(X_n, y|\Theta) := -\log P_n(y|X_n, \Theta)$,
which we use in training our (C)DNNs.
Given the network parameters $\Theta$,
this adversary can compute the derivative of this loss function 
by backpropagating derivatives through the network.
This suggests that our adversary can accomplish its goal by searching for a new input sequence $\hat{X}$ via gradient descent on the loss function with any label $y$ that differs from the ground truth. This is the approach used by Szegedy et al. \cite{Szegedy2014} 
in the context of image object recognition. 

A local minimum of \eqref{eq:opt-adverse} can be found using projected gradient descent, 
initialised with the exemplar $\hat{X}^{(0)} \leftarrow X$, and iterating
\begin{equation}
\hat{X}^{(k+1)} \leftarrow \mathcal{P}_\mathcal{C}(\hat{X}^{(k)} + \mu\nabla \mathcal{L}(\hat{X}^{(k)}, y|\Theta)) 
\end{equation}
where the scalar $\mu$ is the gradient descent step size,
and $\mathcal{P}_\mathcal{C}(\cdot)$ computes the least squares projection of its argument onto the set $\mathcal{C}(X)$ defined in (\ref{eq:feasset}). 
Note that we define operations on sequences element-wise, e.g., $\nabla \mathcal{L}(\hat{X}^{(k)}, y|\Theta) = (\nabla \mathcal{L}(\hat{X}_n^{(k)}, y|\Theta))_{n=0}^{N-1}$. 

  \begin{algorithm}[t]
    \caption{From exemplar sequence $X$
    search for adversarial sequence $\hat{X}$ with maximal perturbation SNR
    in at most $k_{\max}$ steps
    that makes a (C)DNN with parameters $\Theta$ apply label $y$ with confidence $R_{\min}$.} 
    \label{alg:proj-grad-stft}
    \begin{algorithmic}[1]
      \STATE \textbf{parameters: } $y$, SNR, $\mu, R_{\min}, \Theta$, $k_{\max}$
      \STATE \textbf{init: } $X^{(0)} = X, k=0$
     \REPEAT        
      \STATE $V \leftarrow X^{(k)} + \mu\nabla \mathcal{L}(X^{(k)},y|\Theta)$ \COMMENT{Gradient step}
      \STATE $W \leftarrow \mathcal{P}_{GL}(\max(0,V))$ \COMMENT{Find valid sequence}
      \STATE $\nu \leftarrow \max(0,\frac{1}{N}\sqrt{\sum_{n=0}^{N-1}||W_n-X_n||_2^2}/\epsilon(\text{SNR}) - 1)$ \COMMENT{Lagrange mult.}
      \STATE $X^{(k+1)} \leftarrow (1+\nu)^{-1}(W + \nu X)$ \COMMENT{SNR constraint}
      \STATE $k \leftarrow k+1$
      \UNTIL{$\frac{1}{N}\sum_{n=0}^{N-1} P_n(y|X_n^{(k+1),\Theta})
      \geq R_{\min}$ or $k = k_{\max}$}  
      \STATE \textbf{return: } $\hat{X} = X^{(k)}$
    \end{algorithmic}
  \end{algorithm}
  
The main difficulty with this approach is that not all sequences $\hat{X}$ can be mapped back to valid time-domain signals $\hat{x}$. This is because the analysis in \eqref{eq:dft} uses overlapping windows, which causes adjacent elements in the sequence $X$ to become dependent. This means that individual elements from the sequence $X$ cannot be adjusted arbitrarily if we want $X$ to have an analog in the time-domain.
Therefore, in order to generate valid adversarial examples, we include an additional processing step that projects the sequence $\hat{X}$ onto the space of time-frequency coefficients arising from valid time-domain sequences. This is done using the Griffin and Lim algorithm \cite{griffin1984}, which seeks to minimise 
\begin{equation}
\mathcal{P}_{GL}(\hat{X}) = \min_{Z\in\mathcal{X}} \sum_{n=0}^{N-1} || Z_n - \hat{X}_n||_2^2
\end{equation}
where $\mathcal{X}=\{X=(X_n)_{n=0}^{N-1} : \mathcal{P}_{GL}(X)=X\}$ denotes the set of all valid sequences. This minimization can be performed using alternating projections, and we have found that in practice  
it is sufficient to apply a single set of projections. We do this by first rebuilding a complex valued time-frequency representation from the sequence $\hat{X}$
\begin{multline}
U[m,u] = \\
\begin{cases}
   \hat{X}_{\lfloor u/T \rfloor}[m,u\bmod T]e^{j\Phi[m,u]} & 0 \leq m < D \\
   \hat{X}_{\lfloor u/T \rfloor}[D-m, u\bmod T]e^{j\Phi[m,u]} & D \leq m < L.
  \end{cases}
\end{multline}
where $D = L/2+1$ and $\Phi[m,u] \triangleq \angle \mathcal{F}(x)$ is the phase from the exemplar's Fourier transform. The inverse Fourier transform $\mathcal{F}^{-1}(U)$ is a time-domain signal, and so the Fourier transform of this signal, $\mathcal{F}\circ\mathcal{F}^{-1}(U)$, will yield a valid DFT spectrum  that can be used to build a valid input sequence for our (C)DNN, i.e., by replacing $\mathcal{F}(x)$ by $\mathcal{F}\circ\mathcal{F}^{-1}(U)$ in \eqref{eq:in_seq}.

The pseudo-code in Alg.~\ref{alg:proj-grad-stft} summarises this approach. 
The algorithm may be terminated 
when the mean posterior of the target adversarial label 
exceeds the threshold $R_{\min}$, 
or after a maximum number of epochs $k_{\max}$ 
(in which case an adversary cannot be found above the minimum SNR).



\subsection{Training with adversaries for music audio}

As per \cite{Szegedy2014} and \cite{Goodfellow2015a}, 
we can attempt to use our adversary 
as a regulariser, and to create systems robust against adversarial inputs. 
In particular, we create adversaries for the (C)DNN discussed above, 
and use them to generate a (possibly) infinite supply of new samples during training. 
The iterative procedure for generating adversaries in Alg~\ref{alg:proj-grad-stft} 
is too slow to be practical for training, which requires on the order of 50 to 200 training epochs. 
Therefore, we apply the single gradient step procedure suggested in \cite{Goodfellow2015a}. 
In our experience, this procedure often generates inputs 
that confuse the network, although not typically with a high confidence. 
The pseudo-code in Alg.~\ref{alg:adv_trainer} illustrates our training algorithm,
where $(\mathbf{X}, \mathbf{Y})$ represent the training data, 
i.e., the set of input audio sequences and their labels,
and $\hat{\mathbf{Y}}$ is a set of adversarial labels.


 \begin{algorithm}[t]
    \caption{Train (C)DNN using database of labeled sequences $(\mathbf{X}, \mathbf{Y})$
    and fast adversarial generation \cite{Goodfellow2015a}, with
    $\varepsilon$ and $\mu$ the gradient descent step sizes 
    for adjusting the adversarial inputs and network weights, respectively.}
    \label{alg:adv_trainer}
    \begin{algorithmic}[1]
      \STATE \textbf{parameters: $\varepsilon$, $\mu$} 
      \STATE \textbf{init: } (C)DNN parameters $\Theta$ to small random weights
     \REPEAT        
     \STATE select $\hat{\mathbf{Y}}$ uniformly $\{1, \ldots, K\}^N$
      \STATE $\hat{\mathbf{X}} \leftarrow \mathbf{X} + \varepsilon\nabla \mathcal{L}(\mathbf{X},\hat{\mathbf{Y}}|\Theta)$ \COMMENT{Generate adversarial ex.}
      \STATE $\Theta \leftarrow \Theta + \mu\nabla \mathcal{L}(\hat{\mathbf{X}}, \mathbf{Y}|\Theta)$ \COMMENT{Model update}
      \UNTIL{Stopping condition}  
    \end{algorithmic}
  \end{algorithm}

\begin{figure*}[t]
  \begin{center}
  \subfigure[{\em GTZAN} fault-filtered: $23.0 \pm 4.5$dB]{
    \includegraphics[width=0.45\linewidth]{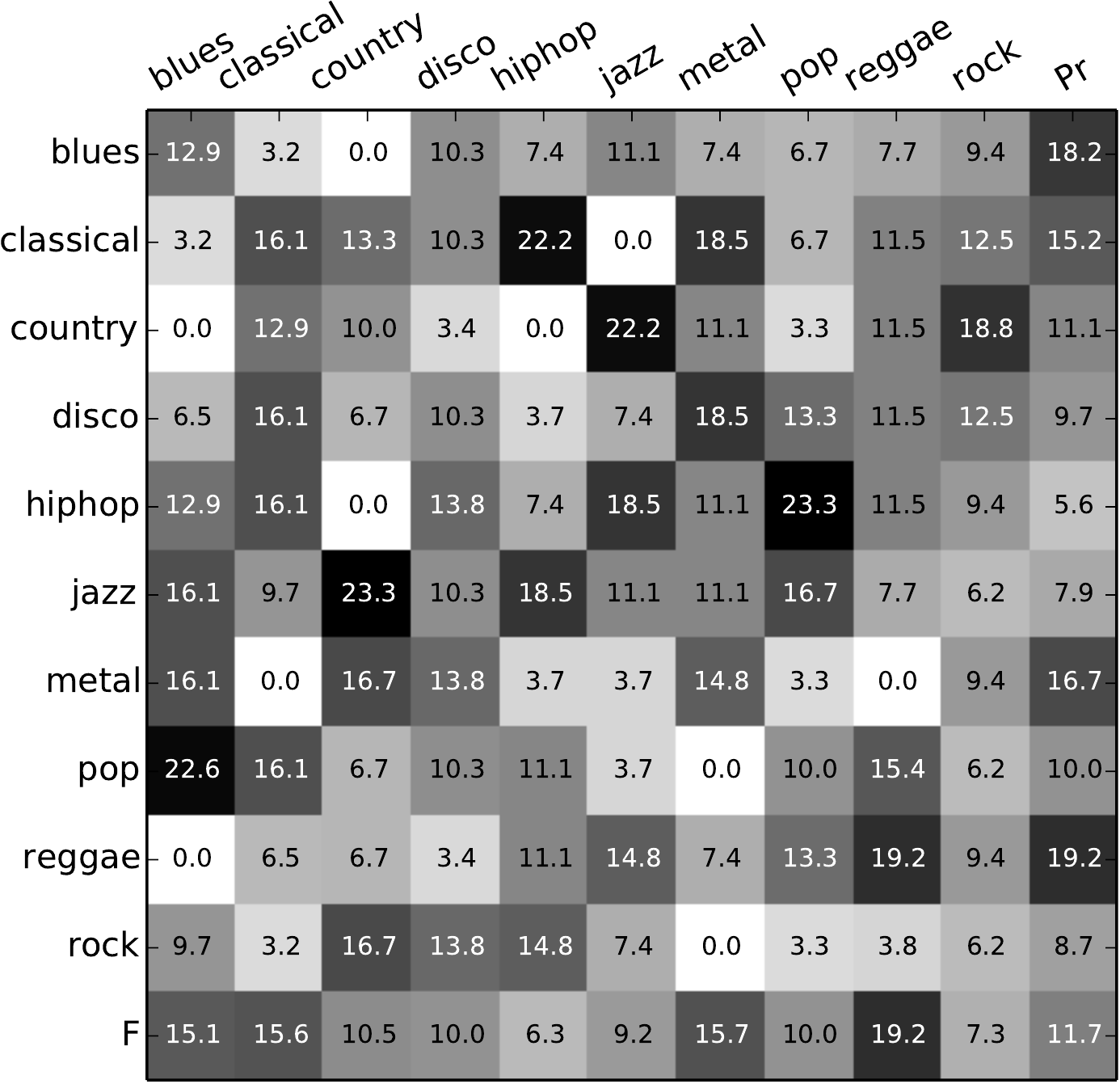}}
    \subfigure[{\em LMD} artist-filtered: $15.78 \pm 4.65$dB]{
    \includegraphics[width=0.45\linewidth]{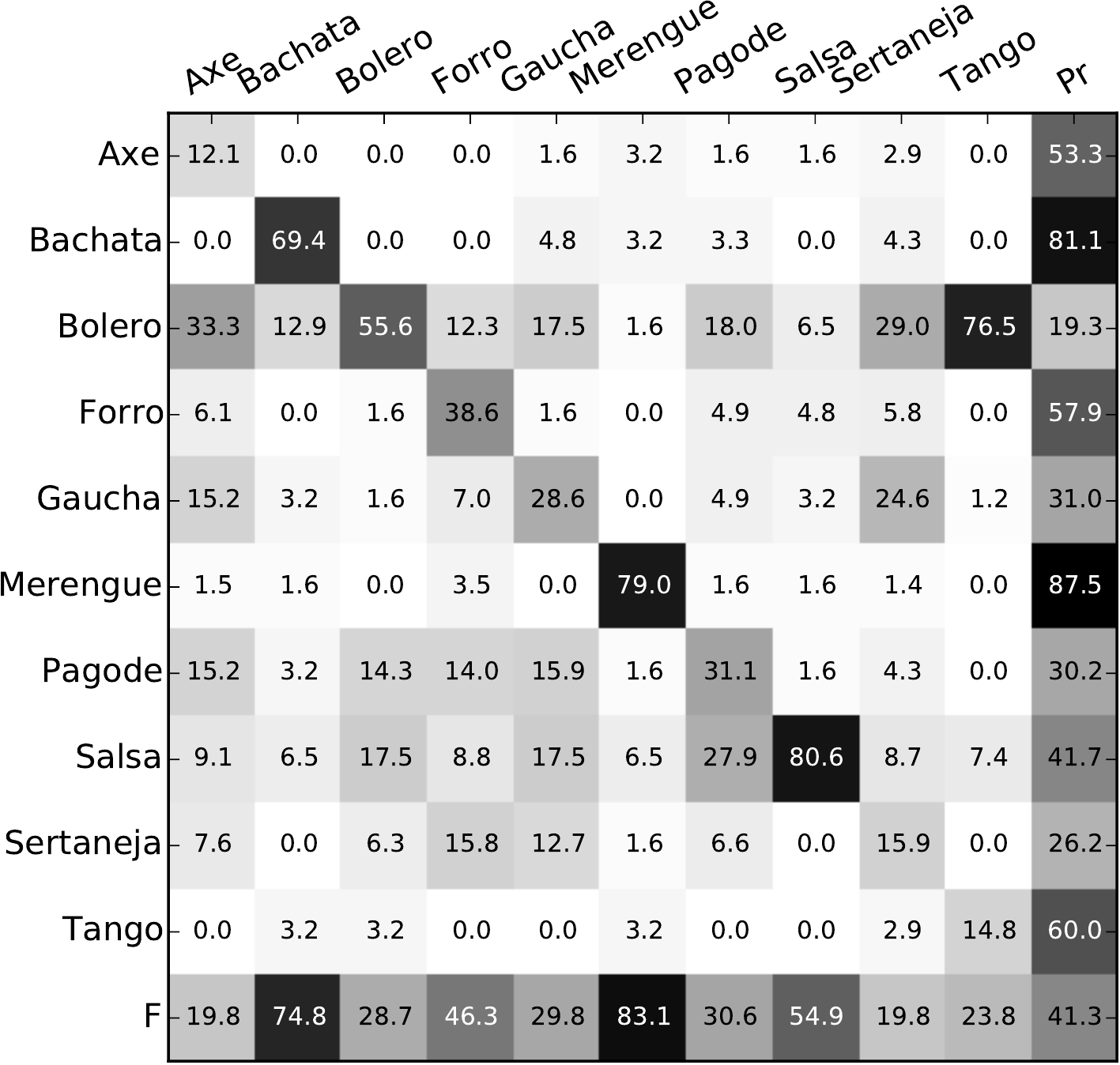}}
  \end{center}
  \caption{For the DNN-based classifier in Fig. \ref{fig:DNN-GTZAN-baselines}(b) and 
  the CDNN in Fig. \ref{fig:deep-LMD-baselines}(d),
but with all input intercepted by an adversary intent on making
the maximum posterior correct with probability $p=0.1$.
For this adversary, $R_{\min}=0.9$, SNR $=15$dB, $\mu=0.1$, 
and $k_{\max} = 100$.
Sub-captions show the resulting SNR (mean $\pm$ standard deviation)
for the adversarial test sets of {\em GTZAN} ($N = 290$)
and {\em LMD} ($N = 646$).}
  \label{fig:CDNNGTZANLMDrandom}
\end{figure*}

\begin{table*}[t]
\centering
{\footnotesize
\begin{tabular}{|r@{\,}|@{\,}c@{\,}|@{\,}c@{\,}|@{\,}c@{\,}|@{\,}c@{\,}|@{\,}c@{\,}|@{\,}c@{\,}|@{\,}c@{\,}|@{\,}c@{\,}|@{\,}c@{\,}|@{\,}c@{\,}|} \hline
& \multicolumn{10}{c@{\,}|}{{\bf Classification in {\em GTZAN}}} \\ 
{\bf Music excerpt} & {\em Blues} & {\em Classical} & {\em Country} & {\em Disco} & {\em Hiphop} & {\em Jazz} & {\em Metal} & {\em Pop} & {\em Reggae} & {\em Rock}\\ \hline 
{\em Little Richard, ``Last Year's Race Horse''} &  32 (23)  &  29 (23) &  36 (25) &  36 (26)  &  36 (25)  &  33 (24) &  32 (24) &  31 (25) &  42 (26) &  36 (25)\\ \hline 
{\em Rossini, ``William Tell Overture''} &  32 (25) &  37 (30) &  40 (29) &  43 (28) &  34 (24) &  36 (29) &  33 (25) &  34 (26) &  37 (26) &  37 (28)\\ \hline 
{\em Willie Nelson, ``A Horse Called Music''} &  25 ( ) &  25 (20) &  30 (27) &  30 (20) &  26 (19) &  30 (25) &  27 (23) &  21 (20) &  30 (23) &  29 (23)\\ \hline 
{\em Simian Mobile Disco, ``10000 Horses Can't Be Wrong''} &  31 (30) &  36 (31) &  38 (32) &  45 (34) &  41 (33) &  40 (32) &  33 (31) &  47 (34) &  42 (33) &  38 (33)\\ \hline 
{\em Rubber Bandits, ``Horse Outside''} &  27 (27) &  27 (27) &  36 (29) &  42 (31) &  38 (29) &  34 (28) &  32 (28) &  37 (29) &  36 (29) &  35 (29)\\ \hline 
{\em Leonard Gaskin, ``Riders in the Sky''} &  32 (23) &  30 (25) &  32 (23) &  35 (25) &  31 (22) &  35 (29) &  34 (23) &  26 (23) &  35 (25) &  35 (24)\\ \hline 
{\em Jethro Tull, ``Heavy Horses''} &  29 (26) &  28 (26) &  40 (29) &  42 (29) &  38 (28) &  36 (28) &  34 (28) &  34 (28) &  37 (28) &  36 (29)\\ \hline 
{\em Echo and The Bunnymen, ``Bring on the Dancing Horses''} &  29 (25) &  28 (26) &  38 (28) &  43 (28) &  35 (26) &  34 (26) &  33 (26) &  33 (26) &  36 (27) &  38 (28)\\ \hline 
{\em Count Prince Miller, ``Mule Train''} &  32 (30) &  29 (30) &  41 (33) &  37 (34) &  43 (33) &  36 (31) &  33 (31) &  42 (34) &  40 (33) &  33 (33)\\ \hline 
{\em Rolling Stones, ``Wild Horses''} &  30 (22) &  32 (24) &  37 (25) &  40 (25) &  31 (22) &  34 (25) &  31 (26) &  32 (23) &  37 (25) &  37 (26)\\ \hline 

\end{tabular}
}
\caption{SNR of perturbations produced by two ensembles of adversaries 
that intercept the input to the system in Fig. \ref{fig:DNN-GTZAN-baselines}(b)
and have it produce all classifications possible with  
confidence thresholds $R_{\min}=0.5$ ($R_{\min}=0.9$ in brackets). 
The average SNR is $34.5$ ($26.8$) dB. 
This table can be heard at \url{http://www.eecs.qmul.ac.uk/~sturm/research/DNN_adversaries}.}
\label{tab:exp2results}\vspace{-0.25in}
\end{table*}

\vspace{-0.1in}
\section{Experimental Results}
\label{sec:experimental_results}
We can design an adversary (Alg. \ref{alg:proj-grad-stft})
such that it will attempt to make a system behave in different ways.
For instance, an adversary could
attempt to perturb an input within some limit (SNR) 
such that the (C)DNN makes a high-confidence classification ($R_{\min} \approx 1$) 
that is correct with probability $p$.
Another adversary could attempt to make the system 
label any input using the same label.
We can also make an ensemble of adversaries
such that they produce adversarial examples
that a (C)DNN classifies in every possible way.

\begin{figure*}[t]
  \begin{center}
    \includegraphics[width=0.95\linewidth]{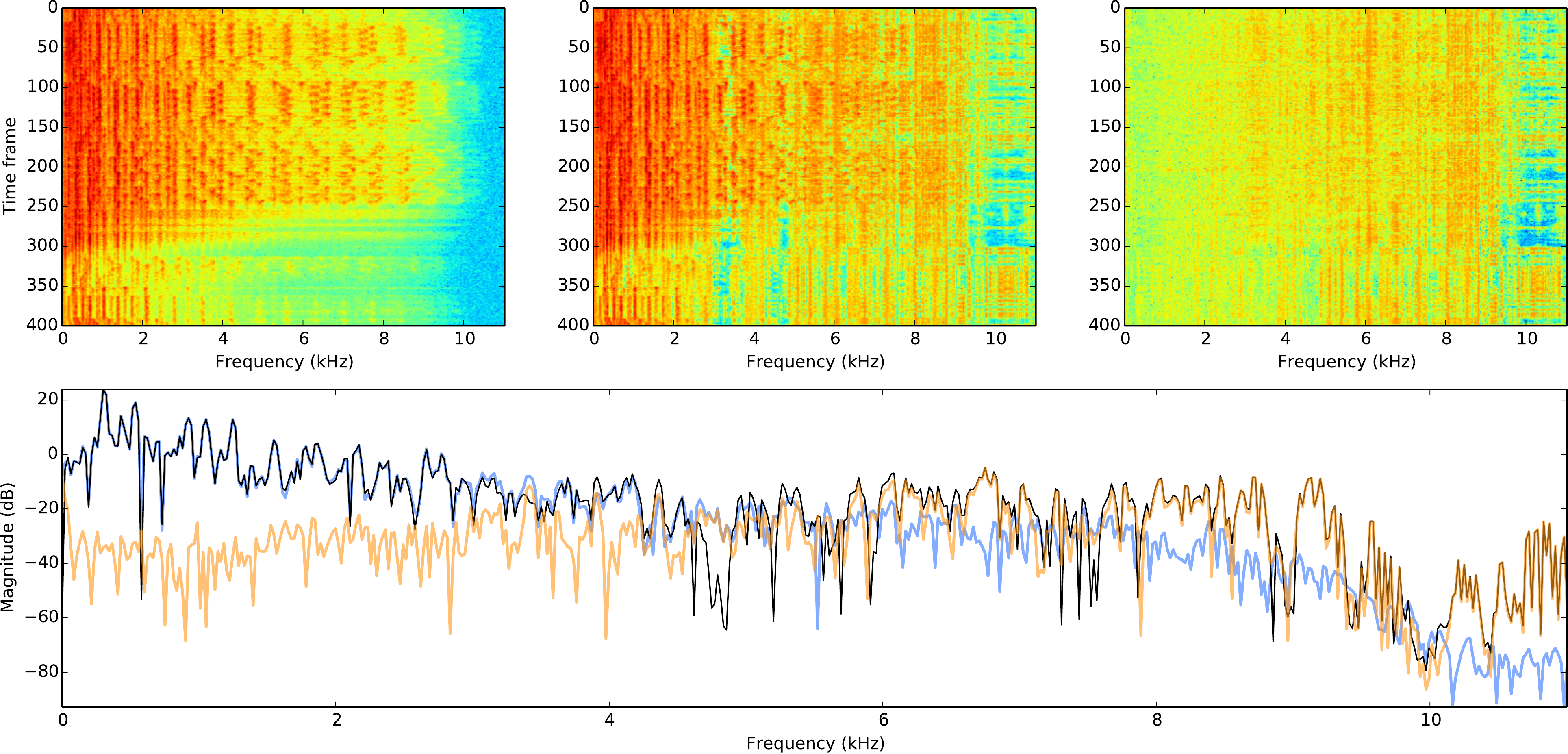}
  \end{center}\vspace{-0.1in}
  \caption{Top left: spectrogram excerpt from {\em GTZAN} Classical ``21''
(Mozart, Symphony No. 39 Finale) that the DNN-based system in
Fig. \ref{fig:DNN-GTZAN-baselines}(b) classifies as {\em Classical}. 
Top middle: spectrogram of adversarial example classified as {\em Reggae}. 
Top right: spectrogram of the difference of the two. 
Bottom: magnitude spectrum of one frame (1024 samples) of the original (light blue), 
adversarial example (black), 
and difference (orange).
Note that all excerpts in {\em GTZAN} have a sampling rate of $22050$ Hz.
The SNR $=21.1$dB.}
  \label{fig:adversary}\vspace{-0.2in}
\end{figure*}

We define our adversaries (Alg. \ref{alg:proj-grad-stft}) using:
$R_{\min}=0.9$, SNR $=15$dB, $\mu=0.1$, and $k_{\max} = 100$,
and with the directive to make the (C)DNN correct with probability $p=0.1$.
More concretely, for each test observations,
the adversary draws uniformly one of the dataset labels $y$,
then seeks to find in no more than $k_{\max} = 100$ iterations
using step size $\mu=0.1$ a valid perturbation  
no larger than $15$dB SNR,
and which the (C)DNN labels as $y$ with confidence $R_{\min}=0.9$.
Figure \ref{fig:CDNNGTZANLMDrandom}(a) shows the FoM of 
the DNN-based classification system in Fig. \ref{fig:DNN-GTZAN-baselines}(b),
but with input intercepted by this adversary.
Note that in this case the classification is performed by
the same random forest classifier using the aggregated
hidden layer activations, but the adversary is unaware of this.
In other words, it is only trying to force the DNN to misclassify
inputs that have been subject to minor perturbations.
Compared with a normalised accuracy of $0.49$ in Fig. \ref{fig:DNN-GTZAN-baselines}(b),
we see our adversary has successfully confused the random forest classifier
to be no better than random.
Figure~\ref{fig:adversary} shows one of the adversarial examples from this experiment.
Apart from some significant high-frequency deviations, 
the spectrum of the adversary is very similar to that of the original.
The SNR in this example is $21.1$dB.

Figure \ref{fig:CDNNGTZANLMDrandom}(b) shows the FoM
of the CDNN classification system in Fig. \ref{fig:deep-LMD-baselines}(d)
attacked by the same adversary.
In this case, the CDNN proved more difficult to fool,
but still the adversary is able to significantly reduce the
normalised classification accuracy from $0.63$ to $0.41$
with high confidence classifications at rather high SNR.
If we reduce the minimum confidence $R_{\min}=0.5$
and lessen the SNR constraint to $-300$ dB,
then the adversary makes the CDNN perform
even worse: a normalised accuracy of $0.28$
with a mean SNR of $11.15 \pm 8.32$ dB.

For the same system in Fig. \ref{fig:DNN-GTZAN-baselines}(b),
and using $R_{\min}=0.9$, SNR $=15$dB, $\mu=0.1$ and $k_{\max} = 100$,
we show in \cite{Kereliuk2015} that we able to create adversaries 
that make the system always right, always wrong, and always select ``Jazz.''
Table \ref{tab:exp2results} shows the results of two ensembles of adversaries,
each intent on making the system in Fig. \ref{fig:DNN-GTZAN-baselines}(b)
choose one of every label in {\em GTZAN} for the same music with 
SNR $=15$dB, $\mu=0.1$ and $k_{\max} = 100$.
The adversaries of one ensemble insist upon a classification confidence
of at least $R_{\min}=0.5$; and in the other of at least $0.9$.
These music recordings are the same 30-second excerpts used in \cite{Sturm2013g}.
We see that in all case by one, 
the ensembles are able to elicit high confidence classifications
from the system with minor perturbations of the input.
We also see that larger perturbations are produced on average
when the adversaries insist on a higher minimum confidence: 
$34.5$ dB for a confidence of at least $R_{\min}=0.5$,
and $26.8$ dB for a confidence of at least $R_{\min}=0.9$.

These results can be heard here: \url{http://www.eecs.qmul.ac.uk/~sturm/research/DNN_adversaries}.
We find that the perturbations caused by these adversaries are certainly perceptible,
unlike those found for image data in \cite{Szegedy2014} and \cite{Goodfellow2015a};
however, the distortion is very minor, and the music remains {\em exactly} the same,
e.g., pitches, rhythm, lyrics, instrumentation, dynamics, and style all remain the same.

\begin{table}[t]
\centering
{\footnotesize
\begin{tabular}{|r@{\,}|@{\,}c@{\,}|@{\,}c@{\,}|@{\,}c@{\,}|} \hline
& {\em Norm.} & {\em Norm. Acc. } & {\em SNR (dB) } \\  
{\bf Deep Learning System} & {\em Acc} & {\em  w/ Adversary} & {\em mean $\pm$ std. dev.} \\ \hline 
DNN-LMD Fig. \ref{fig:deep-LMD-baselines}(b) & 0.63 & 0.03 & 37.8$\pm$4.6 \\ \hline
DNN-LMD+ADV & 0.55 & 0.06 & 36.5$\pm$5.4 \\ \hline
CDNN-LMD Fig. \ref{fig:deep-LMD-baselines}(d) & 0.63 & 0.21 & 9.62$\pm$5.8 \\ \hline
CDNN-LMD+ADV & 0.56 & 0.21 & 9.74$\pm$6.4 \\ \hline
\end{tabular}
}
\caption{Results of applying adversary to make 
systems in Fig. \ref{fig:deep-LMD-baselines}(b,d) always incorrect,
and after training with adversary (Alg. \ref{alg:adv_trainer}).}
\label{tab:exp4results}\vspace{-0.35in}
\end{table}

We now perform an experiment to compare (C)DNNs 
trained with adversarial examples (as per Alg. \ref{alg:adv_trainer}) 
to the systems in Fig. \ref{fig:deep-LMD-baselines}(b,d). 
To do this we test the response of these systems against an adversary 
aimed at {\em always} eliciting an incorrect response.
(This is different from the adversary used above,
which seeks to make the system correct with probability $p=0.1$.)
For this experiment, we set $R_{\min} = 0.5$ and SNR to $-300$ dB
in order to allow arbitrarily large perturbations to force misclassifications. 
Table \ref{tab:exp4results} illustrates the results of this experiment from which we observe several interesting results. Column 1 shows the normalized accuracy on the original test set (with no adversary present). We see that training against adversarial examples leads to a slight deflation in accuracy on new test data. Column 2 shows the normalized accuracy of these systems against our adversary intent on forcing a 100\% error rate. We see that the CDNN systems are more robust to this adversary, and that the systems trained against adversarial examples confer little to no advantage. Column 3 shows the average perturbation size of the adversarial examples that led to misclassifications. We notice that larger perturbations (corresponding to lower SNRs) were required to get the CDNN systems to misclassify test inputs. 
The minimum SNR produced was $0.11$ dB, while the maximum was $47.6$ dB.
The results of this experiment point to the conclusions that a) the CDNN systems are more robust to this adversary; and b) training against adversarial examples (contrary to what we hypothesized) does not seem reduce the misclassification rate against new adversarial examples. A possible explanation for the latter results is that, due to the high-dimensional nature of the input space, the set of possible adversarial examples is densely packed, so that training on a small number of these points is not sufficient to allow the systems to generalize to new adversarial examples.

\vspace{-0.1in}
\section{Discussion}
\label{sec:discussion}
Returning to the broadest question motivating our work,
we seek to measure the contribution of
deep learning to music content analysis.
The previous sections describe a series of experiments
we have conducted using deep learning systems of a variety of architectures,
which we have trained and tested in two different partitions of 
two benchmark music datasets
We have evaluated the robustness of these systems to an adversary
that has complete knowledge of the classifiers,
and have also investigated the use of an adversary in the training
of deep learning systems.

\begin{figure*}[t]
\centering\hspace{-0.3in}
\subfigure[{\em GTZAN} fault-filtered]{
\includegraphics[width=0.45\linewidth]{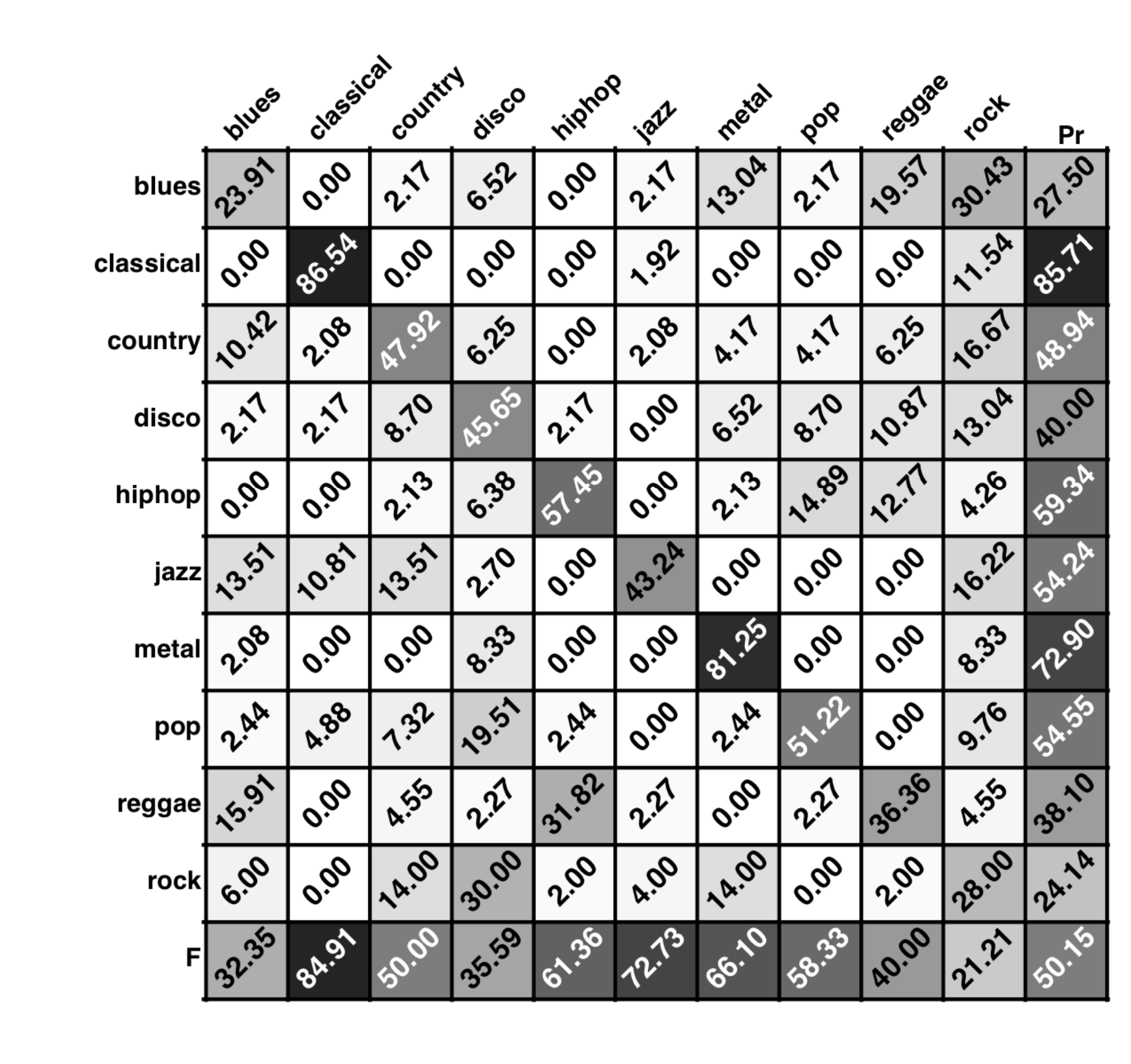}}\hspace{-0.15in}
\subfigure[{\em LMD} artist-filtered]{
\includegraphics[width=0.45\linewidth]{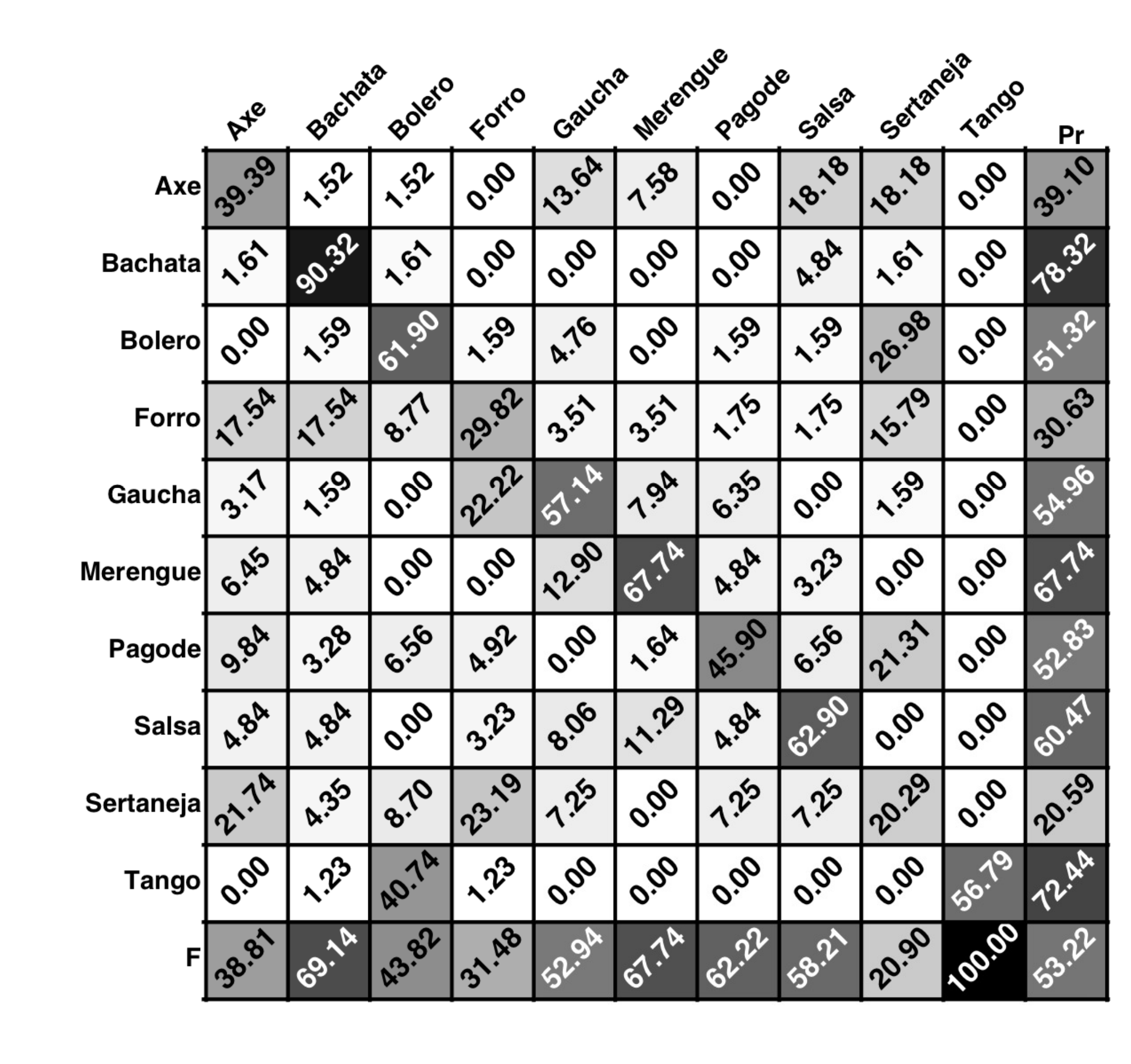}}
\caption{As in Fig. \ref{fig:DNN-GTZAN-baselines},
FoM for majority vote of minimum Mahalanobis distance classification
of mean and variances over 5-second ``texture'' windows 
of zero-crossings and the first 13 MFCCs 
computed from 46 ms windows hopped 50\%.
}
\label{fig:FoM_handcrafted}\vspace{-0.15in}
\end{figure*}

Our experimental results in Fig. \ref{fig:DNN-GTZAN-baselines}
and Table \ref{tab:SigtiaTable} are essentially
reproductions of those reported in \cite{Sigtia2014}.
Based on the results of their experiments with random partitionings of {\em GTZAN},
Sigtia et al. \cite{Sigtia2014} claim that
their DNN-based systems learn features that ``better represent the audio''
than standard or ``hand-crafted'' features, e.g., 
those referenced in \cite{Henaff2011} like MFCCs.
Similar conclusions are made about the deep learning systems in \cite{Hamel2010},
also based on experiments using a random partitioning of {\em GTZAN}.
However, we see in Fig. \ref{fig:DNN-GTZAN-baselines} and 
Table \ref{tab:SigtiaTable} that when we consider the faults
in the {\em GTZAN} dataset and partition it along artist lines, 
as for the {\em LMD} dataset in Fig. \ref{fig:deep-LMD-baselines},
our deep learning systems perform significantly worse. 
This is an expected outcome \cite{Pampalk2005b,Flexer2010,Sturm2013h},
but the artist information in {\em GTZAN} was not available until 2012 \cite{Sturm2012b}.

This motivates the question of whether DNN-based systems 
really do perform better than that of a classifier using standard, low-level and ``hand-crafted'' features.
To examine this, we build baseline systems that use low-level features,
and train and test them in the same fault-filtered partition of {\em GTZAN} as in Fig. \ref{fig:DNN-GTZAN-baselines}(b),
and the artist-filtered partition of {\em LMD} as in Fig. \ref{fig:deep-LMD-baselines}(b,d).
Mimicking \cite{Hamel2010,Sigtia2014}, we compute these features based on 
a short-time analysis using 46ms frames hopped by 50\%. 
From each frame we extract the first 13 Mel-frequency cepstral coefficients (MFCCs) 
and zero-crossings, and compute their mean and variance 
over five-second texture windows (which are also hopped by 50\%). 
We combine the features of the training and validation sets
of the fault-filtered partition of {\em GTZAN},
and the artist filtered partition of {\em LMD}.
Both systems use a minimum Mahalanobis distance classifier,
and assign a class by majority vote 
from the classifications of the individual texture windows. 
Figure \ref{fig:FoM_handcrafted} shows the FoM produced by 
these baseline systems.
We see that for {\em GTZAN} it actually reproduces more ground truth 
than the DNN in Fig. \ref{fig:DNN-GTZAN-baselines}(b)
and all but one in Table \ref{tab:SigtiaTable}.
Our simple baseline system for {\em LMD} reproduces 
much less ground truth than the (C)DNN in Fig. \ref{fig:deep-LMD-baselines}(b,d).
Nonetheless, we have no reason to accept the conclusion 
that deep learning features ``perform better'' than ``hand-crafted'' features 
for the particular architectures considered here and those in \cite{Hamel2010,Sigtia2014}.
Different experiments are needed to address such a conclusion.

A tempting conclusion is that since the normalised classification accuracies
in Figs. \ref{fig:DNN-GTZAN-baselines}(b) and \ref{fig:deep-LMD-baselines}(d)
are extremely unlikely to arise by chance 
($p<10^{-62}$ for {\em GTZAN}
and $p<10^{-290}$ for {\em LMD} by a Binomial test)
it is therefore entirely reasonable to reject the hypothesis that 
our (C)DNN are choosing outputs at random.
Hence, one might argue that
these (C)DNN must have learned features that are ``relevant'' to
music genre recognition \cite{Hamel2010,Sigtia2014,Humphrey2013}.
This argument appears throughout the MIR research discipline \cite{Sturm2013h},
and turns on the strong assumption that
there are only two ways 
a system can reproduce the ground truth of a dataset:
by chance or by learning to solve a specific problem
thought to be well-posed by a cleanly labeled dataset \cite{Sturm2015b}.
In fact, there is a third way a system can reproduce
the ground truth of a music dataset: by learning to exploit
characteristics shared between the training and testing datasets
that arise not from a relationship in the real world,
but from the curation and partitioning of a dataset
in the experimental design of an evaluation
\cite{Sturm2013g,Sturm2013h,Sturm2014}.
Since the evaluations producing  
Figs. \ref{fig:DNN-GTZAN-baselines} and \ref{fig:deep-LMD-baselines},
as well as all results in \cite{Hamel2010,Sigtia2014},
not to mention a significant number of published studies in MIR \cite{Sturm2013h},
do not control for this third way, we cannot validly conclude upon the ``relevance''
of whatever has been learned by these music content analysis systems.

A notion of this problem is given by the significant decreases in the FoM we measure 
when partitioning {\em GTZAN} and {\em LMD} along artist lines.
By doing so, we are controlling for {\em some} independent variables 
that a system might be exploiting to reproduce ground truth,
but which arguably have little relevance to the high-level labels of the dataset \cite{Sturm2013h}.
More concretely, consider that all 100 excerpts labeled {\em Pop}
in {\em GTZAN} come from recordings of music by four artists, 25 from each artist.
If we train and test a system on a random partition of {\em GTZAN},
we cannot know whether the system is recognising {\em Pop},
recognising the artist, or recognising other aspects
that may or may not be related to {\em Pop}.
If we train a system instead with {\em Pop} excerpts by three artists,
test with the {\em Pop} excerpts by the fourth artist, 
then we might be testing something closer to {\em Pop} recognition.
This all depends on defining what knowledge is relevant to the problem.


A common retort to these arguments is that 
a system should be able to reproduce ground truth ``by any means.''
One thereby defines ``relevant knowledge'' as {\em any} correlations
that helps a system reproduce an amount of dataset ground truth 
that is inconsistent with chance.
However, this can lead to circular reasoning:
system X has learned ``relevant knowledge''
because it reproduces Y amount of ground truth;
system X reproduces Y amount of ground truth
because it has learned ``relevant knowledge.''
It is also deaf to one of the major aims of research in 
music content analysis \cite{Casey2008a}:
``to make music, or information about music, easier to find.''
If a music content analysis system is describing music
in ways that do not align with those of its users,
then its usability is in jeopardy 
no matter its FoM in benchmark datasets \cite{Urbano2013,Schedl2013}.
Finally, this means that the problem
thought to be well-posed by a cleanly labeled dataset
can be many things simultaneously ---
which leads to the problem of how to validly compare
apples and oranges \cite{Sturm2015b}.
In other words, why compare systems
when they are {\em solving} different problems?
This also applies to the comparisons above
with the FoM in Fig. \ref{fig:FoM_handcrafted}.


While we have no idea whether our (C)DNN systems
in Fig. \ref{fig:deep-LMD-baselines} are exploiting ``irrelevant'' characteristics in {\em LMD}, 
our experimental results with adversaries 
in Figs. \ref{fig:CDNNGTZANLMDrandom} and \ref{fig:adversary}, 
and Tables \ref{tab:exp2results} and \ref{tab:exp4results},
indicate that their decision machinery is incredibly sensitive
in very strange ways.
Our adversaries are able to fool the high-performing deep learning systems
by perturbing their input in minor ways.
Auditioning the results in Table \ref{tab:exp2results}
show that while the music in each recording remains exactly the same,
and the perturbations are very small,
the DNN is nearly always fooled into choosing with high confidence
every class it has supposedly learned.
The CDNN is similarly defeated by our adversary;
however, it is quite notable that it requires perturbations
of far lower SNR than does the DNN.
We are currently studying the reasons for this.

Our application of adversaries here is close to 
the ``method of irrelevant transformations''
that we apply in \cite{Sturm2013g,Sturm2014,Sturm2015a}
to assess the internal models of music content analysis systems,
and to test the hypothesis, ``the system is using relevant criteria to make its decisions.''
In \cite{Sturm2013g}, we take a brute force approach
whereby we apply random but linear time-invariant
and minor filtering to inputs of systems
trained in three different music recording datasets 
until their FoM becomes perfect or random.
We also make each system apply every one of its classes
to the same music recordings in Table \ref{tab:exp2results}.\footnote{These results 
can be auditioned here: \url{http://www.eecs.qmul.ac.uk/~sturm/research/TM_expt2/index.html}}
In \cite{Sturm2014}, we instead apply 
subtle pitch-preserving time-stretching of music recordings
to fool a deep learning system trained  
in the benchmark music dataset {\em BALLROOM} \cite{Dixon2004}.
We find that through such a transformation we can make 
the system perform perfectly or 
no better than random by applying tempo changes of at most 6\%
to test dataset recordings.
We find a similar result for the same kind of deep learning system 
but trained in {\em LMD} \cite{Sturm2015a}.

Our adversary in Alg. \ref{alg:proj-grad-stft} moves instead 
right to the achilles heel of a deep learning system,
coaxing it to behave in arbitrary ways for an input
simply by making minor perturbations to the sampled audio waveform 
that have no effect on the music content it possesses.
We observe in Fig. \ref{fig:adversary} and auditioning Table \ref{tab:exp2results}
that the low- to mid-frequency content of adversarial examples 
differs very little from the original recordings,
but find more significant differences in the high-frequency spectra.
This suggests that the distribution of energy in the high-frequency spectrum 
has significant impact on the decision machinery of our (C)DNN.
The apparent high relevance of such slight characteristics in proportion
to that of the actual musical content of a music recording
does not bode well for one of the most important aims of machine learning: {\em generalisation}.

As observed by Goodfellow et al. \cite{Goodfellow2015a}
in their deep learning systems taught to recognise objects in images,
the impressive FoM we measure of our 
deep learning systems may be merely a colourful ``Potemkin village.'' 
Employing an adversary to scratch a little below the surface 
reveals the FoM to be curiously hollow.
A system that appears to be solving a complex problem
but actually is not is what we term a ``horse'' \cite{Sturm2013g},
which is a nod to the famous horse Clever Hans:
a real horse that {\em appeared} to be a capable mathematician
but was merely responding to involuntary cues
that went undetected because his public demonstrations
had no validity to attest to such an ability.
Measuring the number of correct answers Hans gives
in an uncontrolled environment 
does not give reason to conclude he comprehends 
what he appears to be doing.
It is the same with the experiments we perform above
with systems labelling observations in {\em GTZAN} and {\em LMD}.
In fact, Goodfellow et al. \cite{Goodfellow2015a} come to the same conclusion:
``The existence of adversarial examples suggests that ... 
being able to correctly label the test data does not imply that our models truly 
understand the tasks we have asked them to perform'' \cite{Goodfellow2015a}.
This observation is now well-known in MIR \cite{Sturm2012e,Sturm2013g,Sturm2013h,Sturm2014d},
but deserves to be repeated.

\section{Conclusion}\label{sec:future}
In this article, we have shown how to 
adapt the adversary of Szegedy et al. \cite{Szegedy2014} 
to work within the context of music content analysis
using deep learning.
We have shown how our adversary is effective
at fooling deep learning systems of different architectures,
trained on different benchmark datasets.
We find our convolutional networks are
more robust against this adversary than
our deep neural networks.
We have also sought to employ the adversary
as part of the training of these systems,
but find it results in systems that remain 
as sensitive to the same adversary.

It is of course not very popular for one to be an ``adversary'' to research,
moving quickly to refute conclusions and break systems reported in the literature;
however, we insist that breaking systems leads ultimately to progress.
Considerable insight can be gained by looking behind 
the veil of performance metrics in an attempt to determine 
the mechanisms by which a system operates,
and whether the evaluation is any valid reflection of 
the qualities we wish to measure.
Such probing is necessary if we are truly interested in ascertaining 
what a system has learned to do, what its vulnerabilities might be, 
how it compares to competing systems supposedly solving the same problem,
and how well we can expect it to perform when used in real-world applications.

\section*{Acknowledgments}
CK and JL were supported in part by the Danish Council for Strategic Research of the
Danish Agency for Science Technology and Innovation under the CoSound project,
case number 11-115328. 
This publication only reflects the authors' views.

\bibliographystyle{plain}
\bibliography{deep,BibAnnon,genre,emotion}

\begin{thebibliography}{10}

\bibitem{Allen1977a}
J.~B. Allen and L.~Rabiner.
\newblock A unified approach to short-time {F}ourier analysis and synthesis.
\newblock {\em Proc. IEEE}, 65(11):1558--1564, Nov. 1977.

\bibitem{Aucouturier2002b}
J.-J. Aucouturier and F.~Pachet.
\newblock Scaling up music playlist generation.
\newblock In {\em Multimedia and Expo, 2002. ICME '02. Proceedings. 2002 IEEE
  International Conference on}, volume~1, pages 105--108 vol.1, 2002.

\bibitem{Aucouturier2004}
J-.J. Aucouturier and F.~Pachet.
\newblock Improving timbre similarity: How high is the sky?
\newblock {\em J. of Negative Results in Speech and Audio Sciences}, 1(1),
  2004.

\bibitem{Bastien2012}
Fr{\'{e}}d{\'{e}}ric Bastien, Pascal Lamblin, Razvan Pascanu, James Bergstra,
  Ian~J. Goodfellow, Arnaud Bergeron, Nicolas Bouchard, and Yoshua Bengio.
\newblock Theano: new features and speed improvements.
\newblock Deep Learning and Unsupervised Feature Learning NIPS 2012 Workshop,
  2012.

\bibitem{Battenberg2012}
E.~Battenberg and D.~Wessel.
\newblock Analyzing drum patterns using conditional deep belief networks.
\newblock In {\em Proc. ISMIR}, 2012.

\bibitem{Bengio2015a}
Y.~Bengio, I.~Goodfellow, and A.~Courville.
\newblock {\em Deep Learning}.
\newblock MIT Press, 2015 (in preparation).

\bibitem{Bengio2009}
Yoshua Bengio.
\newblock {Learning deep architectures for AI}.
\newblock {\em Foundations and trends in Machine Learning}, 2(1):1--127, 2009.

\bibitem{Bergstra2010}
James Bergstra, Olivier Breuleux, Fr{\'{e}}d{\'{e}}ric Bastien, Pascal Lamblin,
  Razvan Pascanu, Guillaume Desjardins, Joseph Turian, David Warde-Farley, and
  Yoshua Bengio.
\newblock Theano: a {CPU} and {GPU} math expression compiler.
\newblock In {\em Proceedings of the Python for Scientific Computing Conference
  ({SciPy})}, June 2010.
\newblock Oral Presentation.

\bibitem{Bertin-Mahieux2010c}
T.~Bertin-Mahieux, D.~Eck, and M.~Mandel.
\newblock Automatic tagging of audio: The state-of-the-art.
\newblock In W.~Wang, editor, {\em Machine Audition: Principles, Algorithms and
  Systems}. IGI Publishing, 2010.

\bibitem{Bertin-Mahieux2011}
T.~Bertin-Mahieux, D.~P.W. Ellis, B.~Whitman, and P.~Lamere.
\newblock The million song dataset.
\newblock In {\em Proc. ISMIR}, 2011.

\bibitem{Boulanger-Lewandowski2013}
N.~Boulanger-Lewandowski, Y.~Bengio, and P.~Vincent.
\newblock Audio chord recognition with recurrent neural networks.
\newblock In {\em Proc. ISMIR}, 2013.

\bibitem{Burges2003}
C.~J.~C. Burges, J.~C. Platt, and S.~Jana.
\newblock Distortion discriminant analysis for audio fingerprinting.
\newblock {\em IEEE Trans. Speech Audio Process.}, 11(3):165--174, May 2003.

\bibitem{Casey2008}
M.~Casey, C.~Rhodes, and M.~Slaney.
\newblock Analysis of minimum distances in high-dimensional musical spaces.
\newblock {\em IEEE Trans. Audio, Speech, Lang. Process.}, 16(5):1015--1028,
  July 2008.

\bibitem{Casey2008a}
M.~Casey, R.~Veltkamp, M.~Goto, M.~Leman, C.~Rhodes, and M.~Slaney.
\newblock Content-based music information retrieval: Current directions and
  future challenges.
\newblock {\em Proc. IEEE}, 96(4):668--696, Apr. 2008.

\bibitem{Collins2010}
Nick Collins.
\newblock Computational analysis of musical influence: A musicological case
  study using mir tools.
\newblock In {\em ISMIR}, pages 177--182, 2010.

\bibitem{Dalvi2004a}
N.~Dalvi, P.~Domingos, Mausam, S.~Sanghai, and D.~Verma.
\newblock Adversarial classification.
\newblock {\em KDD}, 2004.

\bibitem{Deng2014}
L.~Deng and D.~Yu.
\newblock {\em Deep Learning: Methods and Applications}.
\newblock Now Publishers, 2014.

\bibitem{Dieleman2011}
S.~Dieleman, P.~Brakel, and B.~Schrauwen.
\newblock Audio-based music classification with a pretrained convolutional
  network.
\newblock In {\em Proc. ISMIR}, 2011.

\bibitem{Dieleman2014}
S.~Dieleman and B.~Schrauwen.
\newblock End-to-end learning for music audio.
\newblock In {\em Acoustics, Speech and Signal Processing (ICASSP), 2014 IEEE
  International Conference on}, pages 6964--6968, May 2014.

\bibitem{Dixon2004}
S.~Dixon, F.~Gouyon, and G.~Widmer.
\newblock Towards characterisation of music via rhythmic patterns.
\newblock In {\em Proc. ISMIR}, pages 509--517, 2004.

\bibitem{Ewert2014}
S.~Ewert, B.~Pardo, M.~Muller, and M.D. Plumbley.
\newblock Score-informed source separation for musical audio recordings: An
  overview.
\newblock {\em Signal Processing Magazine, IEEE}, 31(3):116--124, May 2014.

\bibitem{Flexer2007}
A.~Flexer.
\newblock A closer look on artist filters for musical genre classification.
\newblock In {\em Proc. ISMIR}, pages 341--344, Sep. 2007.

\bibitem{Flexer2010}
A.~Flexer, D.~Schnitzer, M.~Gasser, and T.~Pohle.
\newblock Combining features reduces hubness in audio similarity.
\newblock In {\em Proc. Int. Symp. Music Info. Retrieval}, 2010.

\bibitem{Goodfellow2015a}
I.~J. Goodfellow, J.~Shlens, and C.~Szegedy.
\newblock Explaining and harnessing adversarial examples.
\newblock In {\em Proc. ICLR}, 2015.

\bibitem{griffin1984}
Daniel Griffin and Jae~S Lim.
\newblock Signal estimation from modified short-time fourier transform.
\newblock {\em Acoustics, Speech and Signal Processing, IEEE Transactions on},
  32(2):236--243, 1984.

\bibitem{Griffith1999}
Niall Griffith and Peter~M Todd.
\newblock {\em Musical networks: Parallel distributed perception and
  performance}.
\newblock MIT Press, 1999.

\bibitem{Gu2014}
S.~{Gu} and L.~{Rigazio}.
\newblock {Towards Deep Neural Network Architectures Robust to Adversarial
  Examples}.
\newblock {\em ArXiv e-prints}, December 2014.

\bibitem{Hamel2010}
P.~Hamel and D.~Eck.
\newblock Learning features from music audio with deep belief networks.
\newblock In {\em Proc. ISMIR}, 2010.

\bibitem{Hastie2009}
T.~Hastie, R.~Tibshirani, and J.~Friedman.
\newblock {\em The Elements of Statistical Learning: Data Mining, Inference,
  and Prediction}.
\newblock Springer-Verlag, 2 edition, 2009.

\bibitem{Henaff2011}
M.~Henaff, K.~Jarrett, K.~Kavukcuoglu, and Y.~LeCun.
\newblock Unsupervised learning of sparse features for scalable audio
  classification.
\newblock In {\em Proc. Int. Soc. Music Info. Retrieval}, Miami, FL, Oct. 2011.

\bibitem{Humphrey2013}
E.~J. Humphrey, J.~P. Bello, and Y.~LeCun.
\newblock Feature learning and deep architectures: New directions for music
  informatics.
\newblock {\em J. Intell. Info. Systems}, 41(3):461--481, 2013.

\bibitem{Humphrey2014}
E.J. Humphrey and J.P. Bello.
\newblock From music audio to chord tablature: Teaching deep convolutional
  networks toplay guitar.
\newblock In {\em Acoustics, Speech and Signal Processing (ICASSP), 2014 IEEE
  International Conference on}, pages 6974--6978, May 2014.

\bibitem{Kereliuk2015}
C.~Kereliuk, B.~L. Sturm, and J.~Larsen.
\newblock Deep learning, audio adversaries, and music content analysis.
\newblock In {\em Proc. WASPAA}, 2015.

\bibitem{Lee2009c}
H.~Lee, Y.~Largman, P.~Pham, and A.~Y. Ng.
\newblock Unsupervised feature learning for audio classification using
  convolutional deep belief networks.
\newblock In {\em Proc. Neural Info. Process. Systems}, Vancouver, B.C.,
  Canada, Dec. 2009.

\bibitem{Li2010}
T.~LH. Li, A.~B. Chan, and A.~HW. Chun.
\newblock Automatic musical pattern feature extraction using convolutional
  neural network.
\newblock In {\em Proc. Int. Conf. Data Mining and Applications}, 2010.

\bibitem{Matityaho1995}
B.~Matityaho and M.~Furst.
\newblock Neural network based model for classification of music type.
\newblock In {\em Proc. Conv. Electrical and Elect. Eng. in Israel}, pages
  1--5, Mar. 1995.

\bibitem{Montavon2012}
G.~Montavon, G.~B. Orr, and K.-R. M\"uller, editors.
\newblock {\em Neural Networks, Tricks of the Trade, Reloaded}.
\newblock Lecture Notes in Computer Science (LNCS 7700). Springer, 2012.

\bibitem{Nguyen2014a}
A.~Nguyen, J.~Yosinski, and J.~Clune.
\newblock Deep neural networks are easily fooled: High confidence predictions
  for unrecognizable images.
\newblock In {\em Proc. NIPS}, 2014.

\bibitem{Pampalk2005b}
E.~Pampalk, A.~Flexer, and G.~Widmer.
\newblock Improvements of audio-based music similarity and genre
  classification.
\newblock In {\em Proc. Int. Soc. Music Info. Retrieval}, pages 628--233, Sep.
  2005.

\bibitem{Papadopoulos1999a}
G.~Papadopoulos and G.~Wiggins.
\newblock Ai methods for algorithmic composition: A survey, a critical view and
  future prospects.
\newblock In {\em Proc. AISB Symposim on Musical Creativity}, pages 110--117,
  1999.

\bibitem{Pikrakis2013}
A.~Pikrakis.
\newblock A deep learning approach to rhythm modeling with applications.
\newblock In {\em Proc. Int. Workshop Machine Learning and Music}, 2013.

\bibitem{Schedl2013}
M.~Schedl, A.~Flexer, and J.~Urbano.
\newblock The neglected user in music information retrieval research.
\newblock {\em J. Intell. Info. Systems}, 41(3):523--539, 2013.

\bibitem{Schwarz2006}
D.~Schwarz.
\newblock Concatenative sound synthesis: The early years.
\newblock {\em J. New Music Research}, 35(1):3--22, Mar. 2006.

\bibitem{Sigtia2014}
S.~Sigtia and S.~Dixon.
\newblock Improved music feature learning with deep neural networks.
\newblock In {\em Acoustics, Speech and Signal Processing (ICASSP), 2014 IEEE
  International Conference on}, pages 6959--6963, May 2014.

\bibitem{Silla2008b}
C.~N. Silla, A.~L. Koerich, and C.~A.~A. Kaestner.
\newblock The {Latin} music database.
\newblock In {\em Proc. ISMIR}, 2008.

\bibitem{Sturm2012b}
B.~L. Sturm.
\newblock An analysis of the {GTZAN} music genre dataset.
\newblock In {\em Proc. ACM MIRUM Workshop}, pages 7--12, Nara, Japan, Nov.
  2012.

\bibitem{Sturm2012e}
B.~L. Sturm.
\newblock Classification accuracy is not enough: On the evaluation of music
  genre recognition systems.
\newblock {\em J. Intell. Info. Systems}, 41(3):371--406, 2013.

\bibitem{Sturm2013g}
B.~L. Sturm.
\newblock A simple method to determine if a music information retrieval system
  is a ``horse''.
\newblock {\em IEEE Trans. Multimedia}, 16(6):1636--1644, 2014.

\bibitem{Sturm2013h}
B.~L. Sturm.
\newblock The state of the art ten years after a state of the art: Future
  research in music information retrieval.
\newblock {\em J. New Music Research}, 43(2):147--172, 2014.

\bibitem{Sturm2014d}
B.~L. Sturm.
\newblock A survey of evaluation in music genre recognition.
\newblock In A.~N\"urnberger, S.~Stober, B.~Larsen, and M.~Detyniecki, editors,
  {\em Adaptive Multimedia Retrieval: Semantics, Context, and Adaptation},
  volume LNCS 8382, pages 29--66, Oct. 2014.

\bibitem{Sturm2015b}
B.~L. Sturm.
\newblock ``horse'' inside: Seeking causes of the behaviours of music content
  analysis systems.
\newblock {\em ACM Computers in Entertainment}, 2015 (submitted).

\bibitem{Sturm2015a}
B.~L. Sturm, C.~Kereliuk, and J.~Larsen.
\newblock ?` el caballo viejo? latin genre recognition with deep learning and
  spectral periodicity.
\newblock In {\em Proc. Int. Conf. on Mathematics and Computation in Music},
  2015.

\bibitem{Sturm2014}
B.~L. Sturm, C.~Kereliuk, and A.~Pikrakis.
\newblock A closer look at deep learning neural networks with low-level
  spectral periodicity features.
\newblock In {\em Proc. Int. Workshop on Cognitive Info. Process.}, 2014.

\bibitem{Szegedy2014}
C.~Szegedy, W.~Zaremba, I.~Sutskever, J.~Bruna, D.~Erhan, I.~Goodfellow, and
  R.~Fergus.
\newblock Intriguing properties of neural networks.
\newblock In {\em Proc. ICLR}, 2014.

\bibitem{Tzanetakis2002}
G.~Tzanetakis and P.~Cook.
\newblock Musical genre classification of audio signals.
\newblock {\em IEEE Trans. Speech Audio Process.}, 10(5):293--302, July 2002.

\bibitem{Urbano2013}
J.~Urbano, M.~Schedl, and X.~Serra.
\newblock Evaluation in music information retrieval.
\newblock {\em J. Intell. Info. Systems}, 41(3):345--369, Dec. 2013.

\bibitem{Oord2013}
A.~van~den Oord, S.~Dieleman, and B.~Schrauwen.
\newblock Deep content-based music recommendation.
\newblock In {\em Proc. NIPS}, 2013.

\bibitem{Vempala2012}
N.~Vempala and F.~Russo.
\newblock Predicting emotion from music audio features using neural networks.
\newblock In {\em Proc. CMMR}, 2012.

\bibitem{Wang2003}
A.~Wang.
\newblock An industrial strength audio search algorithm.
\newblock In {\em Proc. Int. Soc. Music Info. Retrieval}, Oct. 2003.

\bibitem{Weninger2014}
F.~Weninger, F.~Eyben, and B.~Schuller.
\newblock On-line continuous-time music mood regression with deep recurrent
  neural networks.
\newblock In {\em Acoustics, Speech and Signal Processing (ICASSP), 2014 IEEE
  International Conference on}, pages 5412--5416, May 2014.

\bibitem{Whitman2001}
B.~Whitman, G.~Flake, and S.~Lawrence.
\newblock Artist detection in music with minnowmatch.
\newblock {\em Proc. IEEE Workshop on Neural Networks for Signal Processing},
  pages 559--568, 2001.

\bibitem{Wiggins2009}
G.~A. Wiggins.
\newblock Semantic gap?? {S}chemantic schmap!! {M}ethodological considerations
  in the scientific study of music.
\newblock In {\em Proc. IEEE Int. Symp. Mulitmedia}, pages 477--482, Dec. 2009.

\bibitem{Yang2011b}
X.~Yang, Q.~Chen, S.~Zhou, and X.~Wang.
\newblock Deep belief networks for automatic music genre classification.
\newblock In {\em Proc. INTERSPEECH}, pages 2433--2436, 2011.

\bibitem{Yang2011f}
Y.-H. Yang and H.~H. Chen.
\newblock {\em Music Emotion Recognition}.
\newblock CRC Press, 2011.

\bibitem{Zhang2014}
Chiyuan Zhang, G.~Evangelopoulos, S.~Voinea, L.~Rosasco, and T.~Poggio.
\newblock A deep representation for invariance and music classification.
\newblock In {\em Acoustics, Speech and Signal Processing (ICASSP), 2014 IEEE
  International Conference on}, pages 6984--6988, May 2014.

\end{thebibliography}

\end{document}